\documentclass[letterpaper]{article}
\usepackage[preprint]{aaai2027}
\usepackage[hyphens]{url}
\usepackage{graphicx}
\usepackage{natbib}
\usepackage{caption}
\usepackage{algorithm}
\usepackage{algorithmic}
\usepackage{booktabs}
\usepackage{amsmath}
\usepackage{amssymb}
\usepackage{amsfonts}
\usepackage{mathtools}
\usepackage{amsthm}
\usepackage{xcolor}
\usepackage{colortbl}
\usepackage{subcaption}
\usepackage{multirow}
\usepackage{tabularx}
\usepackage{listings}
\usepackage{placeins}


\newcolumntype{C}{>{\centering\arraybackslash}X}
\DeclareMathOperator*{\argmax}{arg\,max}

\theoremstyle{plain}
\newtheorem{theorem}{Theorem}[section]

\newtheorem{lemma}[theorem]{Lemma}

\theoremstyle{definition}

\newtheorem{assumption}[theorem]{Assumption}
\theoremstyle{remark}

\definecolor{grimip_color}{HTML}{4965B0}
\definecolor{smac_color}{HTML}{FBDA83}
\newcommand{\coloredline}[1]{\textcolor{#1}{\rule[0.5ex]{1.5em}{2pt}}}
\newcommand{\promptheading}[1]{%
  \par\addvspace{6pt}%
  \noindent\textbf{#1}\par\nobreak
  \vspace{2pt}%
}
\DeclareFontShape{OT1}{cmtt}{b}{n}{<->cmtt10}{}
\DeclareFontFamily{TS1}{cmtt}{}
\DeclareFontShape{TS1}{cmtt}{m}{n}{<->cmtt10}{}
\DeclareFontShape{TS1}{cmtt}{b}{n}{<->cmtt10}{}
\DeclareFontFamily{OMS}{cmtt}{}
\DeclareFontShape{OMS}{cmtt}{m}{n}{<->ssub*cmsy/m/n}{}
\renewcommand{\ttfamily}{\fontencoding{OT1}\fontfamily{cmtt}\selectfont}
\newcommand{\grimiptitle}{\texttt{GRIMIP}}
\newcommand{\authorblock}[1]{\makebox[0pt][c]{\parbox[t]{\textwidth}{\centering\fontsize{9.5pt}{11pt}\selectfont #1}}}
\newcommand{\emailblock}[1]{\parbox[t]{0.92\textwidth}{\centering\texttt{#1}}}
\lstdefinestyle{prompt}{
  basicstyle=\ttfamily\fontsize{6pt}{6.5pt}\selectfont,
  breaklines=true,
  breakatwhitespace=false,
  columns=fullflexible,
  keepspaces=true,
  showstringspaces=false,
  frame=single,
  framerule=0.3pt,
  xleftmargin=1pt,
  xrightmargin=1pt,
  aboveskip=4pt,
  belowskip=4pt,
  literate={_}{{\char95}}1 {→}{{$\rightarrow$}}1
}

\title{\grimiptitle: A General Framework for Instance-Specific\\[-0.2ex]
MIP Solver Configuration Using Large Language Models}
\author{%
  \authorblock{Yidong Luo\textsuperscript{1},
  Chenguang Wang\textsuperscript{1},
  Fangzhou Zhu\textsuperscript{2},
  Tao Zhong\textsuperscript{2},
  Tianshu Yu\textsuperscript{1}\thanks{Corresponding author.}} \\
  \normalfont
  \textsuperscript{1}School of Data Science, The Chinese University of Hong Kong, Shenzhen \\
  \textsuperscript{2}Noah's Ark Lab, Huawei \\
  \emailblock{\{yidongluo,chenguangwang\}@link.cuhk.edu.cn, yutianshu@cuhk.edu.cn, \{zhufangzhou,zhongtao\}@huawei.com} \\
}
\affiliations{}

\begin{document}

\maketitle

\begin{abstract}
Configuring the hyperparameters of Mixed-integer programming (MIP) solvers is a high-dimensional, instance-dependent optimization problem where suboptimal settings can degrade solving time by orders of magnitude. Default configurations are often suboptimal, while traditional tuning methods either suffer from the ``cold-start'' problem and inefficient search or heavily rely on expert experience. This paper introduces \textbf{GRIMIP} (\textbf{\underline{G}}eneral \textbf{\underline{R}}easoning for \textbf{\underline{I}}nstance-specific \textbf{\underline{MIP}} configuration), a novel hybrid intelligence framework that synergistically integrates the semantic reasoning capabilities of Large Language Models (LLMs) with the sample-efficient search of Bayesian Optimization (BO).
GRIMIP is a complete LLM-driven BO framework that applies semantic reasoning throughout space selection, initialization, candidate generation, and feedback-driven refinement, enabling rapid attainment of high-quality configurations with only a small number of expensive solver evaluations.
On seven benchmarks including MIPLIB, GRIMIP achieves over 40\% reduction in Primal-Dual Integral on hard instances, outperforming SMAC and other LLM-assisted BO methods. By granting LLMs sufficient autonomy, GRIMIP combines the expert-level reasoning of LLMs with the efficient search of BO, achieving state-of-the-art performance.
\end{abstract}

\section{Introduction}

Mixed-integer programming (MIP) is a widely used optimization framework for solving various important real-world applications, such as supply chain management~\citep{paschos2014applications}, production planning~\citep{junger2009fifty} and scheduling~\citep{chen2010integrated}. Solving general MIPs is NP-hard because the problem class includes NP-complete 0--1 integer programs~\citep{karp1972reducibility}. The efficacy of solving these complex models hinges on the performance of sophisticated MIP solvers, which is, in turn, critically dependent on their internal hyperparameter configurations \citep{schede2022survey, pan2024beyond}. The default settings of these solvers are often suboptimal for challenging, specific problem instances. Given that modern solvers can feature massive interacting parameters, an exhaustive search for the optimal configuration is computationally prohibitive. This creates a pressing need within both academia and industry for systematic and efficient methods to tune solver parameters for individual, challenging problem instances. The task of tuning these parameters is thus often framed as the optimization of an expensive black-box function.

The need to tune these expensive black-box functions has spurred the development of various automated algorithm configuration methods. While earlier works relied on a variety of traditional heuristics, recent studies have shifted towards learning-based approaches. These methods leverage diverse techniques ranging from mathematical programming~\citep{Iommazzo_2020} and graph representation learning~\citep{valentin2022instancewise} to clustering strategies~\citep{song2023instance_specific} and specialized component optimization~\citep{liu2024l2pmip}. What's more, a particularly prominent and sample-efficient methodology is Bayesian Optimization (BO) \citep{mockus1998application,kushner1964new, jones1998efficient, shahriari2015taking,garnett2023bayesian}, which is well-suited for tuning complex algorithms like MIP solvers. Powerful BO frameworks, such as SMAC~\citep{hutter2011sequential, lindauer2022smac3} and HEBO~\citep{cowen2022hebo}, serve as versatile optimizers capable of operating at varying granularities. They can be employed to find a robust ``one-size-fits-all'' configuration for a population of instances~\citep{hutter2010automated}, or adapted to optimize configurations for individual problem instances. To explicitly address the heterogeneity of instances, the paradigm of per-instance algorithm configuration (PIAC)~\citep{hutter2006performance, leyton2002learning, hutter2014algorithm, xu2008satzilla, belkhir2016feature} has been extensively studied, exemplified by feature-based clustering methods like ISAC~\citep{kadioglu2010isac} and portfolio approaches like Hydra~\citep{xu2010hydra}. Population-based recommendation relies on judging instance similarity, but reliably assessing such similarity remains a highly challenging problem~\citep{luo2026genbench}. However, regardless of whether one employs a population-based or per-instance strategy, two fundamental challenges persist across these paradigms:

\begin{enumerate}

    \item \textbf{Search inefficiency.} Traditional surrogate models struggle to navigate the complex, high-dimensional performance landscape within a limited budget of expensive solver evaluations, often converging to suboptimal configurations.

    \item \textbf{Search space definition.} Defining an effective search space remains a significant barrier: expert-designed spaces risk omitting impactful parameters, while the full parameter space suffers from the curse of dimensionality.

\end{enumerate}

Fundamentally, these methods lack semantic understanding of the problem---a capability that could inform both space selection and search guidance.

Large Language Models (LLMs)~\citep{brown2020language, openai2023gpt4, guo2025deepseekr1} offer a promising avenue to address this gap. 
Their capacity for semantic reasoning enables them to interpret problem structure, relate instance features to solver behavior, and transfer optimization knowledge across domains---capabilities that traditional surrogate models fundamentally lack.
Early attempts leveraged LLMs for zero-shot optimization via direct prediction or code generation~\citep{kochnev2025optuna, zhang2023using}, but these approaches suffer from over-exploitation and entrapment in local optima due to the absence of principled exploration~\citep{sun2025llm_mab}.

Direct one-shot parameter recommendations from LLMs are inconsistent across models and offer no reliable improvement, as shown in Table~\ref{tab:direct_llm_intro}, motivating a closed-loop search that can use feedback to refine configurations and balance exploration with exploitation.

\begin{table}[t]
\centering
\small
\resizebox{\columnwidth}{!}{%
\begin{tabular}{lccc}
\toprule
Method & Mean PDI $\downarrow$ & $\Delta$ vs. Default & Win Rate \\
\midrule
Default & 194.95 & -- & -- \\
DeepSeek-V3.2-Exp & 198.90 & $+2.0\%$ & 48\% \\
GPT-5.5-medium & 201.89 & $+3.6\%$ & 44\% \\
Claude-Opus-4-7 & 192.80 & $-1.1\%$ & 55\% \\
\bottomrule
\end{tabular}
}
\caption{One-shot direct recommendation on IP (100 instances; PDI at 300~s; lower is better). Win rate is measured against the default configuration.}
\label{tab:direct_llm_intro}
\end{table}

Recent work has shifted toward hybrid paradigms that combine LLM reasoning with the theoretical guarantees of BO.
Frameworks such as LLAMBO~\citep{liu2024llmbo} and BO-ICL~\citep{ramos2023bayesian} employ LLMs for warm-starting or as surrogate models, while SLLMBO~\citep{mahammadli2024sequential} integrates LLM suggestions with TPE sampling.
However, these methods position the LLM as an \emph{auxiliary component} within conventional BO pipelines---we term this paradigm \textbf{LLM-assisted BO}.
By constraining the LLM to isolated subtasks (e.g., initialization or mean prediction), they fail to fully exploit its reasoning capabilities and remain bottlenecked by the limitations of traditional surrogates, particularly in quantifying uncertainty for effective exploration-exploitation trade-offs.

In this work, we pioneer the integration of the LLM4BO paradigm into the task of MIP solver tuning. We introduce a complete LLM-driven BO framework that applies semantic reasoning throughout space selection, initialization, candidate generation, and feedback-driven refinement, enabling rapid attainment of high-quality configurations with only a small number of expensive solver evaluations. To address the inherent curse of dimensionality, we incorporate an Automated Space Selection (ASS) module, which leverages the LLM's reasoning to intelligently prune the high-dimensional parameter set into a focused, instance-specific search space. Furthermore, we implement an effective Warm-Starting (WS) mechanism that generates an informed initial portfolio based on structural problem properties. This strategy effectively resolves the ``cold-start" problem, enabling our method to achieve rapid convergence in the early iterations. Through autonomous exploration and continuous reward feedback from the solver, the LLM dynamically updates its prior beliefs over the parameter space. This iterative refinement enables the model to rapidly map and comprehend the overall performance landscape. Ultimately, this synergistic approach achieves significant performance improvements, exceeding 40\% reductions in PDI compared to baseline methods.

We summarize our contributions as follows: 
\begin{enumerate} \item \textbf{Behavioral Analysis:} We conduct a detailed analysis of LLM-driven BO search behavior and identify the root causes of its rapid convergence, showing how early exploitation and diminishing exploration concentrate the search around promising configurations. \item \textbf{Technical:} We introduce three synergistic components: (i) Automated Space Selection for instance-specific search space pruning, (ii) Warm-Starting for informed initialization, and (iii) joint mean-variance prediction for uncertainty-aware acquisition. \item \textbf{Empirical:} Extensive experiments on seven benchmarks demonstrate state-of-the-art performance, with over 40\% PDI reduction on hard datasets (MIRP, MIPLIB) and consistent improvements across 65--99\% of instances. \item \textbf{Practical:} By encapsulating domain expertise within LLM reasoning, GRIMIP lowers the barrier to expert-level solver tuning, enabling practitioners without deep optimization knowledge to achieve competitive configurations. \end{enumerate}

\section{Preliminaries}

\subsection{Mixed-Integer Programming}

MIP is a widely used mathematical optimization framework that involves making decisions under a set of constraints, where some decision variables are required to take integer values. A MIP instance (specifically in its canonical linear form) can be expressed as:
\[
\min \left\{ \mathbf{c}^\top \mathbf{x} : \mathbf{A} \mathbf{x} \leq \mathbf{b}, \mathbf{x} \in \mathbb{Z}^p \times \mathbb{R}^{n-p} \right\}
\]

Here, $\mathbf{x}$ is a vector of $n$ decision variables, of which the first $p$ are restricted to be integers ($\mathbb{Z}^p$) and the remaining $n-p$ are continuous ($\mathbb{R}^{n-p}$). The objective is to minimize the linear function $\mathbf{c}^\top \mathbf{x}$ subject to $m$ linear constraints defined by the coefficient matrix $\mathbf{A} \in \mathbb{R}^{m \times n}$ and the right-hand side vector $\mathbf{b} \in \mathbb{R}^m$.

\subsection{Bayesian Optimization for Hyperparameter Tuning}
\label{subsec:bo_hpo}

We frame the task of finding the optimal configuration for a given problem instance as a complex black-box optimization problem.

Let the set of all possible hyperparameter configurations for a solver be the $d$-dimensional space $\mathcal{X}$. 
A specific configuration is denoted by a vector $c \in \mathcal{X}$, and its performance is measured by a scalar value $y \in \mathbb{R}$. 
The objective is to find the configuration $c^*$ that minimizes a performance function $g$:
\[
c^* = \arg\min_{c \in \mathcal{X}} g(c)
\]
The function $g$ is treated as a ``black box'' because its analytical form is unknown, its evaluations are computationally demanding, and its gradients are inaccessible. 
Given these properties, BO is a highly suitable, sample-efficient methodology. 
Instead of directly optimizing $g$, BO builds a probabilistic model to approximate it and guides the search for the optimum. 
This process involves several key components:

\begin{itemize}
    \item \textbf{Cold Start}:
    The initial set of evaluated points, which forms the first dataset $\mathcal{D}_n$, significantly influences the BO process. A well-chosen initial set can accelerate convergence by providing the surrogate model with a more informative initial representation of the objective function's landscape.
    
    \item \textbf{Probabilistic Surrogate Model}:
    This is the core of BO.
    The surrogate model learns the relationship between hyperparameter configurations and their performance scores from a set of previously observed data points,
    $\mathcal{D}_n = \{(c_i, y_i)\}_{i=1}^n$.
    It produces a \emph{predictive distribution} over the performance $y$ for any configuration $c$:
    \[
    p(y \mid c, \mathcal{D}_n) = \int_{\Lambda} p(y \mid c, \lambda, \mathcal{D}_n)\, p(\lambda \mid \mathcal{D}_n)\, d\lambda
    \]
    In this formulation, the integration is performed over $\lambda$, a latent variable that encapsulates our beliefs about the function's underlying structure.
    The term $p(\lambda \mid \mathcal{D}_n) \propto p(\mathcal{D}_n \mid \lambda)\, p(\lambda)$ represents the posterior distribution of this variable, which is updated from our initial \emph{prior beliefs}, $p(\lambda)$.
    The central challenge, therefore, is to accurately learn this predictive distribution from limited data while incorporating meaningful prior knowledge.
    
    \item \textbf{Candidate Sampling}:
    A sampler proposes a set of new candidate points $\{\tilde{c}_k\}_{k=1}^K$ from the search space $\mathcal{X}$ that are promising for evaluation by the acquisition function.
    
    \item \textbf{Acquisition Function}:
    An acquisition function, such as the Lower Confidence Bound (LCB), evaluates the utility of each candidate point based on the surrogate model's predictive distribution.
    It intelligently balances \textit{exploitation }(choosing points in regions predicted to have good performance) and \textit{exploration} (choosing points in regions with high uncertainty).

\end{itemize}

\section{Methodology} \label{method}

\begin{figure*}[t]
    \centering
    \includegraphics[width=\textwidth]{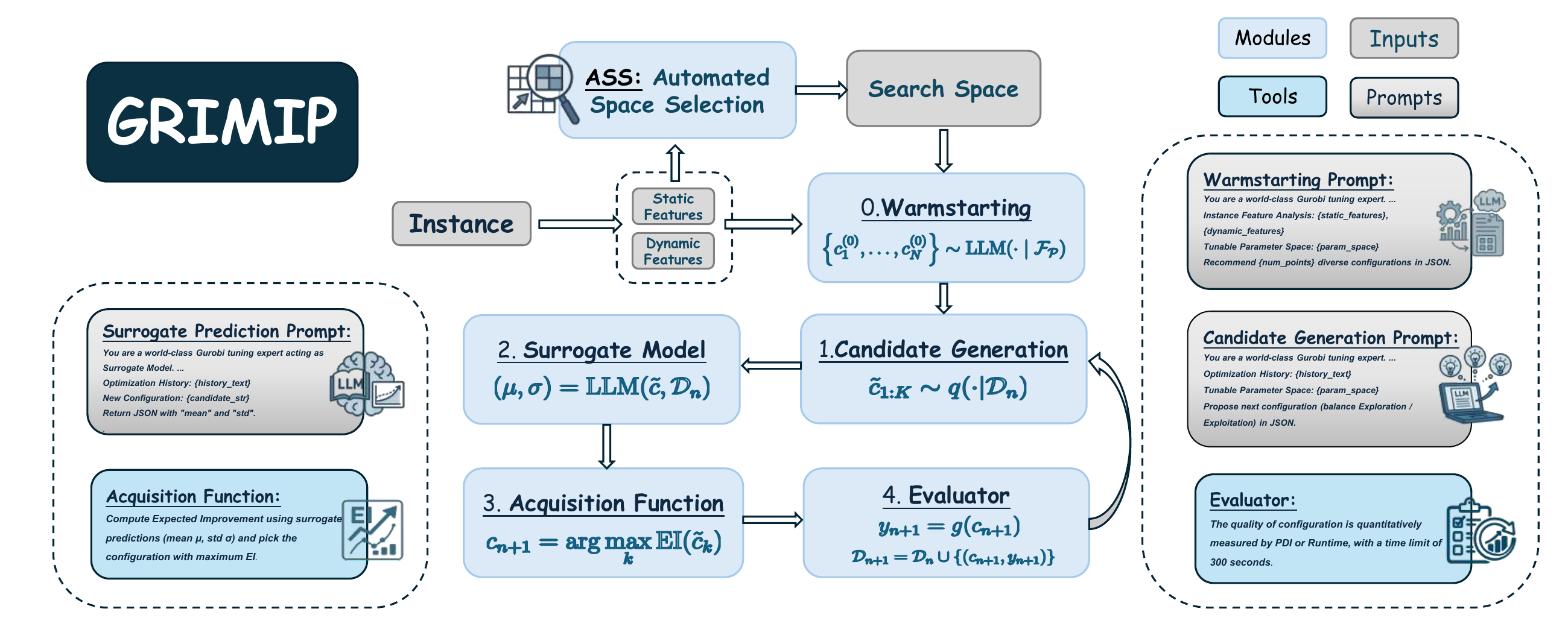}
    \caption{Overview of GRIMIP: This framework utilizes LLM-driven candidate generation, surrogate model prediction, and an Expected Improvement acquisition function to automatically recommend optimal hyperparameter configurations for a given optimization instance.}
    \label{fig:framework}
\end{figure*}

The proposed GRIMIP framework establishes a methodological synergy between the semantic reasoning capabilities of LLMs and the sequential decision-making structure of BO. By embedding the LLM as a core probabilistic component within the optimization loop, our method constructs a hybrid pipeline designed to leverage high-level pattern recognition for efficient MIP solver configuration. As illustrated in Figure~\ref{fig:framework}, the GRIMIP workflow is organized into three coherent stages to systematically navigate the configuration landscape.

\subsection{Automated Search Space Selection (ASS)}

The Gurobi solver provides over 200 tunable parameters (e.g., \texttt{MIPFocus}, \texttt{Heuristics}, \texttt{Cuts}), constituting a high-dimensional search space $\mathcal{X}$. Direct optimization within this entire space is highly inefficient and heavily susceptible to the ``curse of dimensionality,'' which severely diffuses the search focus. To mitigate this issue, our workflow introduces an Automated Space Selection (ASS) module to intelligently prune this space. This module is activated immediately following a 30-second exploratory run, utilizing both static and dynamic instance features $\mathcal{F}_P$, which are extracted in the same manner as in the Warm-Starting (WS) module. We supply the LLM with these features $\mathcal{F}_P$, alongside a comprehensive list of tunable Gurobi parameters, their semantic descriptions (e.g., ``\texttt{MIPFocus}: MIP solution strategy''), and their default ranges. By analyzing $\mathcal{F}_P$, the LLM identifies a low-dimensional parameter subset (restricted to $k=6$ parameters in our default configuration) that it predicts will have the most significant impact on the specific instance. 

Furthermore, our framework supports a \emph{dynamic ASS} mode. When enabled, the LLM is queried after each optimization iteration and provided with the cumulative optimization history $\mathcal{D}_t$. Its task is to periodically re-evaluate the search space, enabling the adaptive inclusion of new parameters deemed influential based on historical performance, and the removal of those exhibiting negligible correlation. This progressively refines the search space as more empirical data is accumulated.

The output of this stage is an instance-specific, reduced-dimensional parameter space $\mathcal{X}_{\text{ASS}} \subset \mathcal{X}$. Ablation studies demonstrate the critical importance of this space-pruning strategy; omitting the ASS module results in the most severe performance degradation among all components. Additionally, a detailed analysis of the rationale behind the LLM's parameter selections is provided in the Configuration Selection Analysis.

\subsection{Warm-starting (WS)}

Unlike traditional methods that start from random points, our workflow begins with an intelligent, instance-specific \emph{warm-starting} stage. For each MIP instance $P$, we first perform a short 30-second preliminary solve using Gurobi's default parameters. This exploratory run allows us to extract a set of key features, $\mathcal{F}_P$, encompassing both \textbf{static features} (e.g., number of variables, types of constraints) and \textbf{dynamic features} (e.g., initial dual gap, number of processed nodes, types of cutting planes applied; the complete feature lists are provided in the supplementary material). These features are then incorporated into a detailed prompt instructing the LLM to act as a Gurobi tuning expert. Based on its analysis of $\mathcal{F}_P$, the LLM generates an initial, diverse population of $N$ high-potential parameter configurations: \(\{c_1^{(0)}, \dots, c_N^{(0)}\} \sim \text{LLM}(\cdot | \mathcal{F}_P).\) This initial population is then thoroughly evaluated against the true performance function $g(\cdot)$ to obtain their corresponding scores, $y_i^{(0)} = g(c_i^{(0)})$. This process creates a high-quality starting dataset $\mathcal{D}_0 = \{ (c_i^{(0)}, y_i^{(0)}) \}_{i=1}^N$ for the subsequent optimization loop, significantly accelerating the overall search process.

\subsection{LLM Probabilistic Surrogate (LLM-PS)} 
After the warm-starting stage, the core of our methodology is an iterative optimization loop in which the LLM fully serves as the \emph{surrogate model}. At the beginning of each iteration $n$, we have a dataset of previously evaluated configurations and their observed performance metrics,\(\mathcal D_n=\{(c_i,y_i)\}_{i=1}^n,\) along with the best performance observed so far, $y^*=\min_{i \le n}y_i$. The loop then unfolds in a sequence of three main steps: Candidate Generation, Performance Prediction, and Acquisition.

\paragraph{Candidate Generation.} Initially, the LLM functions as a candidate sampler. By processing the complete optimization history $\mathcal{D}_n$, the LLM is prompted to propose a novel set of $K$ candidate configurations that exhibit the highest potential for exploration or exploitation.
$$
\tilde c_{1,n+1},\dots,\tilde c_{K, n+1} \sim q(\cdot\,|\,\mathcal D_n).
$$
where $q(\cdot\,|\,\mathcal D_n)$ is the conditional sampling distribution implicitly modeled by the LLM based on the historical data and heuristics learned from its pre-training.

\paragraph{Performance Prediction with Uncertainty.}
Next, the LLM directly acts as the surrogate model. For each proposed candidate configuration $\tilde c_k$, the LLM is prompted to predict both its expected performance (mean $\mu_k$) and the uncertainty of that prediction (standard deviation $\sigma_k$). This is a critical step where the LLM simulates a probabilistic model:
$$
(\tilde \mu_{k,n+1},\tilde \sigma_{k,n+1})=\mathrm{LLM}\!\left(\tilde c_{k,n+1},\mathcal D_n\right),
\qquad k=1,\dots,K.
$$

\paragraph{Acquisition Function and Point Selection.}
Finally, using the predicted mean and uncertainty, we compute a classical acquisition function to select the single best candidate for evaluation. We use the \emph{Expected Improvement} (EI) function, which balances \emph{exploitation} (favoring candidates with low predicted means) and \emph{exploration} (favoring candidates with high predictive uncertainty). For a minimization problem, EI is calculated as:

\[
\begin{aligned}
    \mathrm{EI}_{k,n+1}(\tilde c_{k,n+1})
    &= (y^* - \tilde\mu_{k,n+1})\,\Phi(Z_{k,n+1}) \\
    &\quad + \tilde\sigma_{k,n+1}\,\phi(Z_{k,n+1}),
\end{aligned}
\]
\[
\text{where } Z_{k,n+1} = \frac{y^* - \tilde\mu_{k,n+1}}{\tilde\sigma_{k,n+1}}.
\]

where $\Phi(\cdot)$ and $\phi(\cdot)$ are the CDF and PDF of the standard normal distribution, respectively.
The next configuration to evaluate, $c_{n+1}$, is chosen by maximizing the acquisition function over the set of candidates:
$$
c_{n+1}=\arg\max_{k=1,\dots,K}\mathrm{EI}(\tilde c_{k,n+1})
$$
The selected configuration $c_{n+1}$ is then evaluated on the true performance function, $y_{n+1} = g(c_{n+1})$, and the new pair is added to the dataset for the next iteration: $\mathcal D_{n+1}=\mathcal D_n\cup\{(c_{n+1},y_{n+1})\}$. This loop continues until either a predefined maximum number of iterations is reached, or when no improvement in the best-observed performance is made for a specified number of consecutive iterations (early stopping).

\section{Experiments}

\subsection{Datasets} \label{datasets}

\begin{table}[t]
    \centering
    \scriptsize
    \begin{tabular}{@{}lccc@{}}
        \toprule
        \textbf{Dataset} & \textbf{Type} & \textbf{Vars} & \textbf{Cons} \\
        \midrule
        MIK\textsuperscript{a} & Medium & [255, 520] & [155, 470] \\
        CORAL\textsuperscript{b} & Medium & [62, 573K] & [6, 550K] \\
        MIRP\textsuperscript{c} & Hard & [2K, 194K] & [1K, 109K] \\
        MIPLIB\textsuperscript{d} & Hard & [71, 1.6M] & [7, 2.9M] \\
        \midrule 
        Item Placement\textsuperscript{e} & Hard & 1,083 & 195 \\
        Load Balancing\textsuperscript{e} & Hard & 61,000 & 64,304 \\
        Anonymous\textsuperscript{e} & Hard & 37,881 & 49,603 \\
        \bottomrule
    \end{tabular}
    \par\raggedright\scriptsize 
    \textsuperscript{a}\citep{atamturk2003facets}; mean: Vars=387, Cons=312. \quad
    \textsuperscript{b}\citep{gomes2008connections}; mean: Vars=19,989, Cons=12,482. \\
    \textsuperscript{c}\citep{papageorgiou2014mirplib}; mean: Vars=34,193, Cons=16,170. \quad
    \textsuperscript{d}\citep{MIPLIB2017}; mean: Vars=116,034, Cons=135,733. \\
    \textsuperscript{e}\citep{gasse2022ml4co}.
    \caption{Summary of MIP Datasets}
    \label{tab:dataset_summary}
\end{table}

We evaluate GRIMIP on seven MIP datasets spanning medium-scale instances (MIK and CORAL), structured hard instances (MIRP), heterogeneous hard instances (MIPLIB), and three ML4CO benchmarks (IP, LB, and Anonymous).
Their sources and structural statistics are summarized in Table~\ref{tab:dataset_summary}.

\subsection{Baseline Methods} \label{sec:baselines}
To comprehensively evaluate efficacy, we compare GRIMIP against the following established baselines:
\textcircled{\scriptsize 1}~\textbf{Default}: The default Gurobi parameter configuration.
\textcircled{\scriptsize 2}~\textbf{SMAC-P} \citep{lindauer2022smac3}: A population-based BO approach that seeks a single robust ``one-size-fits-all'' configuration by minimizing the average PDI across random instance subsets.
\textcircled{\scriptsize 3}~\textbf{Random Search}: A per-instance baseline that evaluates $N$ randomly generated configurations, where $N$ matches the total evaluations performed by GRIMIP to ensure a fair budget comparison.
\textcircled{\scriptsize 4}~\textbf{SMAC-I}: A strong per-instance baseline applying SMAC individually to each problem. Its total tuning time is strictly limited to match GRIMIP's wall-clock time for a fair comparison.
\textcircled{\scriptsize 5}~\textbf{GPTT} \citep{gurobi2024}: Gurobi's native tuning tool.
\textcircled{\scriptsize 6}~\textbf{LLAMBO} \citep{liu2024llmbo}: An LLM-driven BO framework for hyperparameter tuning.
\textcircled{\scriptsize 7}~\textbf{ifBO} \citep{rakotoarison2024ifbo}: A BO baseline that uses a deep surrogate (pre-trained transformer) as a prior over learning curves.
\textcircled{\scriptsize 8}~\textbf{TuRBO} \citep{eriksson2019turbo}: A BO baseline based on trust regions, maintaining local GP surrogates inside shrinking/expanding regions.
Further implementation details are provided in the supplementary material.

\subsection{Evaluation Metric}
All of our evaluation experiments were conducted under a strict single-thread constraint (thread = 1) to ensure fair and reproducible comparisons. The complete hardware and software setup is provided in the supplementary material. Performance is assessed using two primary metrics: the \textbf{Primal-Dual Integral} (PDI) and \textbf{Runtime}. The PDI quantifies convergence speed by integrating the primal-dual gap over time within a given time limit, offering a nuanced view of progress on unsolved instances, while Runtime measures the total time to reach \emph{optimality}. Since the MIK and CORAL instance sets are relatively simple and most instances can quickly reach optimality, we adopt Runtime as the primary evaluation metric for these datasets (see Table~\ref{tab:performance_summary}). For the final evaluation, we conduct five in-distribution assessments where the parameter configurations obtained from all methods are benchmarked on the original, complete set of instances. Additional metric details are provided in the supplementary material.

\subsection{Main Results}

\begin{table*}[!t]
\centering
\footnotesize
\setlength{\tabcolsep}{2pt} 

{\scriptsize
\renewcommand{\arraystretch}{0.92}
\setlength{\tabcolsep}{2pt}
\begin{tabularx}{\textwidth}{@{}l *{9}{C}@{}}
\toprule
& \multicolumn{3}{c}{\textbf{Item Placement}~\citep{gasse2022ml4co}} & \multicolumn{3}{c}{\textbf{Load Balancing}~\citep{gasse2022ml4co}} & \multicolumn{3}{c}{\textbf{Anonymous}~\citep{gasse2022ml4co}} \\
\cmidrule(lr){2-4} \cmidrule(lr){5-7} \cmidrule(lr){8-10}
\textbf{Method} & PDI $\downarrow$ & Imp. $\uparrow$ & Eff. $\uparrow$ & PDI $\downarrow$ & Imp. $\uparrow$ & Eff. $\uparrow$ & PDI $\downarrow$ & Imp. $\uparrow$ & Eff. $\uparrow$ \\
\midrule
Default & 194.95 & - & - & 2.72 & - & - & 73.25 & - & - \\
\cmidrule(r){1-1} \cmidrule(lr){2-10}
Random & 194.96 & 0.0\% & 48.6\% & 6.00 & -120.6\% & 64.6\% & 79.63 & -8.7\% & 42.2\% \\
SMAC-P~\citep{lindauer2022smac3} & 161.26 & 17.3\% & 90.6\% & 10.12 & -272.1\% & 76.6\% & 67.80 & 7.4\% & 56.1\% \\
SMAC-I & 152.28 & 21.9\% & 91.6\% & 1.77 & 34.9\% & 76.8\% & 56.22 & 23.3\% & 69.0\% \\
GPTT~\citep{gurobi2024}   & 158.48 & 18.7\% & 55.0\% & 3.11 & -14.3\% & 48.2\% & 72.63 & 0.8\% & 42.9\% \\
LLAMBO~\citep{liu2024llmbo} & 153.29 & 21.4\% & 97.0\% & 1.50 & 44.9\% & 82.2\% & 59.82 & 18.3\% & 73.6\% \\
ifBO~\citep{rakotoarison2024ifbo}   & 147.91 & 24.1\% & 97.0\% & 5.84 & -114.5\% & 10.4\% & 68.73 & 6.2\% & 53.1\% \\
TuRBO~\citep{eriksson2019turbo}  & 149.56 & 23.3\% & 95.0\% & 6.40 & -135.1\% & 10.0\% & 66.55 & 9.2\% & 51.0\% \\
\rowcolor{grimip_color!20}[0pt][0pt] GRIMIP & \textbf{\underline{133.20}} & \textbf{31.7\%} & \textbf{99.0\%} & \textbf{\underline{1.48}} & \textbf{45.6\%} & \textbf{84.2\%} & \textbf{\underline{54.67}} & \textbf{25.4\%} & \textbf{80.6\%} \\
\bottomrule
\end{tabularx}
}

{\scriptsize
\renewcommand{\arraystretch}{0.92}
\setlength{\tabcolsep}{1pt}
\begin{tabularx}{\textwidth}{@{}l *{12}{C}@{}}
\toprule
& \multicolumn{3}{c}{\textbf{MIK}~\citep{atamturk2003facets}} & \multicolumn{3}{c}{\textbf{CORAL}~(Gomes et al., 2008)} & \multicolumn{3}{c}{\textbf{MIRP}~\citep{papageorgiou2014mirplib}} & \multicolumn{3}{c}{\textbf{MIPLIB}~\citep{MIPLIB2017}} \\
\cmidrule(lr){2-4} \cmidrule(lr){5-7} \cmidrule(lr){8-10} \cmidrule(lr){11-13}
\textbf{Method} & \mbox{Time (s) $\downarrow$} & Imp. $\uparrow$ & Eff. $\uparrow$ & \mbox{Time (s) $\downarrow$} & Imp. $\uparrow$ & Eff. $\uparrow$ & PDI $\downarrow$ & Imp. $\uparrow$ & Eff. $\uparrow$ & PDI $\downarrow$ & Imp. $\uparrow$ & Eff. $\uparrow$ \\
\midrule
Default & \textbf{0.204} & - & - & \textbf{99.6} & - & - & 277.24 & - & - & 197.85 & - & - \\
\cmidrule(r){1-1} \cmidrule(lr){2-13}
Random & 0.306 & -50.4\% & 22.2\% & 106.0 & -6.5\% & 56.6\% & 205.69 & 25.8\% & 55.8\% & 178.21 & 9.9\% & 65.9\% \\
SMAC-P & 0.211 & -3.7\% & 21.1\% & 145.0 & -45.7\% & 13.1\% & 208.32 & 24.9\% & 64.8\% & 203.00 & -2.6\% & 24.8\% \\
SMAC-I & 0.211 & -3.5\% & \textbf{40.0\%}& 113.1 & -13.6\% & 66.7\% & 198.76 & 28.3\% & 65.8\% & 189.53 & 4.2\% & 53.5\% \\
GPTT   & 0.208 & -2.2\% & 12.2\% & 100.1 & \textbf{-0.5\%} & 81.6\% & 271.85 & 1.9\% & 38.0\% & 191.32 & 3.3\% & 58.9\% \\
LLAMBO & 0.217 & -6.6\% & 26.7\% & 103.3 & -3.7\% & 69.6\% & 203.89 & 26.5\% & 69.0\% & 179.06 & 11.0\% & 62.0\% \\
ifBO   & 0.435 & -113.5\% & 15.6\% & 102.9 & -3.4\% & 67.4\% & 200.92 & 27.5\% & 73.0\% & 173.86 & 12.1\% & 63.6\% \\
TuRBO  & 3.160 & -1451\% & 28.9\% & 107.6 & -8.1\% & 69.4\% & 195.33 & 29.5\% & 71.0\% & 173.84 & 12.1\% & 63.0\% \\
\rowcolor{grimip_color!20}[0pt][0pt] GRIMIP & \underline{0.205} & \textbf{-0.7\%} & 38.9\% & \underline{100.2} & -0.6\% & \textbf{75.5\%} & \textbf{\underline{159.99}} & \textbf{42.3\%} & \textbf{97.4\%} & \textbf{\underline{172.69}} & \textbf{12.7\%} & \textbf{69.0\%} \\
\bottomrule
\end{tabularx}
}
\caption{Comparison of different tuning methods across datasets. For each dataset, we report the average PDI (lower is better), the percentage improvement over the default configuration (Imp., higher is better), and the percentage of instances where the method outperformed the default (Eff., higher is better).}
\label{tab:performance_summary}
\end{table*}

\paragraph{Performance on Hard Datasets.}
The results summarized in Table~\ref{tab:performance_summary} demonstrate that GRIMIP consistently outperforms all baselines, achieving PDI reductions between \textbf{12.7\% and 45.59\%} on the five most challenging datasets (\textit{IP, LB, A, MIRP, MIPLIB}). This confirms the framework's effectiveness on hard problems, which are the primary targets for instance-specific tuning.

\begin{itemize}
    \item \textbf{Surprising Performance on MIPLIB:} On the prestigious \textit{MIPLIB} benchmark---widely considered the gold standard for solver evaluation---GRIMIP achieved a \textbf{12.72\%} PDI reduction. Improving upon the highly-tuned default settings of Gurobi on this specific dataset is notoriously difficult, underscoring the significance of GRIMIP's gains.
    
    \item \textbf{Significant PDI Reduction:} On \textit{IP} and \textit{LB}, GRIMIP achieved substantial improvements of \textbf{31.67\%} and \textbf{45.59\%} over default settings, respectively. It consistently outperformed SMAC-I, securing the lowest average PDI across both sets.
    
    \item \textbf{Robustness:} The framework exhibited exceptional stability, outperforming default configurations on \textbf{84\% to 99\%} of instances across these datasets.
\end{itemize}

\paragraph{On Medium Datasets:} These simpler datasets were evaluated by total runtime. For simple instances like these, the default parameters are already highly optimized and sufficient, making further improvements difficult to achieve. Critically, the time required for the parameter tuning process itself is many times greater than the few seconds (or less) it takes to solve these instances. In any practical setting, instance-specific tuning would not be applied to problems this simple, as the cost of tuning outweighs any potential benefit. We included these datasets to test the framework's behavior, but as expected, our attempts did not yield significant performance gains.

\begin{figure*}[ht]
    \centering
    \includegraphics[width=\textwidth]{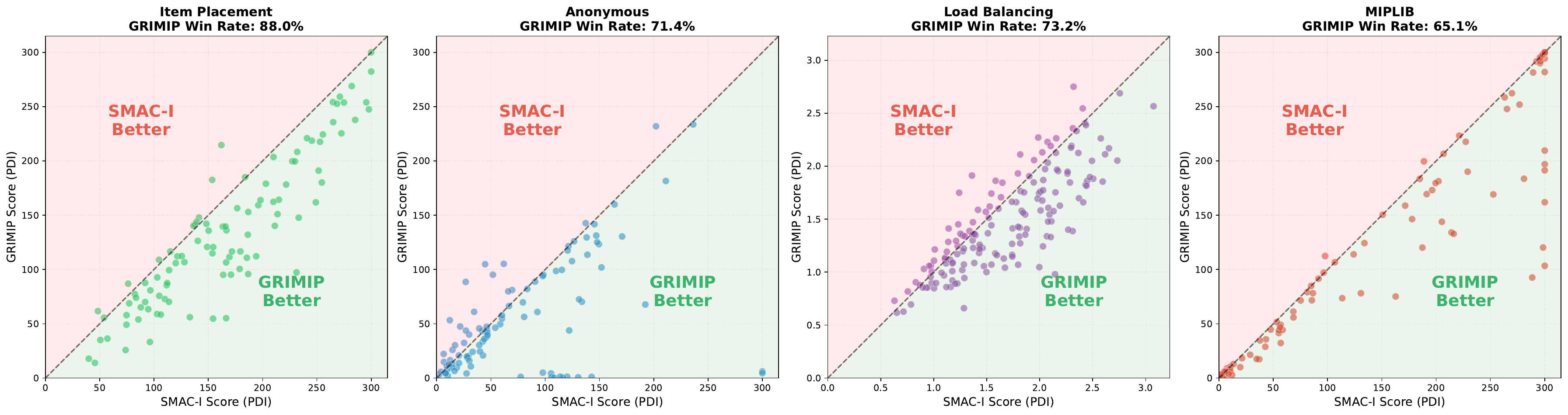}
    \caption{GRIMIP vs SMAC-I PDI Comparison}
    \label{fig:grimip_vs_smaci}
\end{figure*}

\paragraph{Comparison with SMAC-I} Figure~\ref{fig:grimip_vs_smaci} visualizes the instance-wise performance gap. GRIMIP demonstrates clear dominance, with points consistently falling below the diagonal. 

\subsection{Ablation Studies} \label{abl}
\begin{table}[h]
    \footnotesize
    \centering
    \setlength{\tabcolsep}{1.5pt}
    \begin{tabular*}{\columnwidth}{@{\extracolsep{\fill}}lcc@{}}
    \toprule
    Method & Mean PDI & Mean Evals. \\
    \midrule
    GRIMIP & \textbf{133.20} & \textbf{11.30} \\
    GRIMIP w/o WS & 140.83 & 11.36 \\
    GRIMIP w/o std & 146.72 & 11.52 \\
    GRIMIP w/o WS\_Dynamic & 138.44 & 11.56 \\
    GRIMIP w/o ASS (fixed 6 params) & 149.45 & 10.59 \\
    GRIMIP w/o ASS (16 params) & 153.98 & 11.68 \\
    SMAC-I+ASS & 149.77 & 11.68 \\
    \bottomrule
    \end{tabular*}
    \caption{IP ablation results with SMAC-I+ASS.}
    \label{tab:method_comparison}
\end{table}
Table~\ref{tab:method_comparison} validates the synergistic contribution of each GRIMIP component. The \textbf{ASS} proves to be the most critical module: replacing instance-specific selection with a fixed six-parameter space raises the mean PDI to \textbf{149.45}, while expanding the non-adaptive space to 16 parameters further raises it to \textbf{153.98}. These results show that neither a rigid small space nor a broader unpruned space substitutes for instance-aware selection under a limited budget. Regarding initialization, the \textbf{WS} phase effectively resolves the cold-start problem, with \textit{dynamic} features providing a distinct advantage over static-only features (\textbf{133.20 vs. 138.44}). This indicates that capturing solver behavior allows the LLM to identify superior initial basins of attraction that static attributes alone cannot reveal. Finally, our \textbf{joint uncertainty prediction} mechanism significantly outperforms the repeated-sampling baseline (w/o std). Beyond merely reducing computational overhead, this result implies that explicitly querying the LLM for confidence yields a cleaner exploration signal for the acquisition function compared to the noisy variance estimates derived from Monte Carlo sampling.

To isolate the effect of ASS, we run standard SMAC-I in the instance-specific spaces selected by ASS under the same setting and budget (\textbf{SMAC-I+ASS}). Its improvement over fixed-space SMAC-I (Table~\ref{tab:performance_summary}) confirms the effectiveness of ASS.

\vspace{-2pt}

\section{Discussion}

\paragraph{Configuration Selection Analysis.} \label{config_selection}
Figure~\ref{fig:configs} summarizes the parameter subsets selected by DeepSeek-V3.2-Exp. Under a fixed tuning budget, ASS consistently retains a small backbone of high-impact parameters while using instance features such as integrality structure, matrix density, relaxation gaps, and cut activity to allocate the remaining slots.
Across datasets, \texttt{MIPFocus}, \texttt{Heuristics}, \texttt{Cuts}, and \texttt{VarBranch} are selected most frequently, whereas low-leverage parameters are routinely pruned. The remaining choices vary by problem type: binary-dominated IP instances favour cut families such as \texttt{GomoryPasses} and \texttt{CliqueCuts}; non-binary LB instances shift toward \texttt{Method} or \texttt{NodeMethod}; and hard MIRP instances emphasize feasibility-oriented parameters such as \texttt{MIPFocus}, \texttt{Heuristics}, \texttt{Cuts}, and \texttt{Presolve}. This pattern suggests that ASS combines a stable default core with instance-specific adjustments rather than relying on a fixed universal subset.
\begin{center}
    \begin{minipage}{\linewidth}
        \centering
        \includegraphics[width=\linewidth]{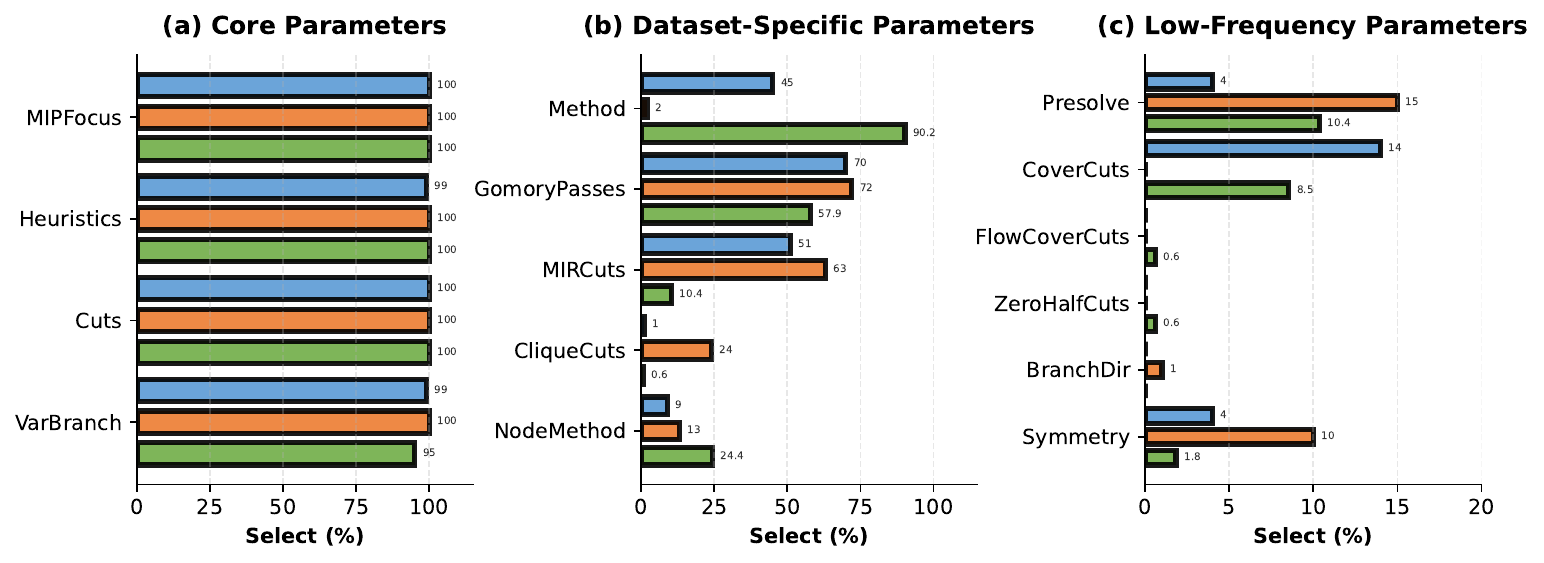}
        \captionof{figure}{Preference of Configuration Selection across Different Datasets.
        Selection frequency (\%) of each parameter across three benchmark datasets:
        \textcolor[HTML]{5B9BD5}{\rule{1.5ex}{1.5ex}} A,
        \textcolor[HTML]{ED7D31}{\rule{1.5ex}{1.5ex}} IP, and
        \textcolor[HTML]{70AD47}{\rule{1.5ex}{1.5ex}} LB.}
        \label{fig:configs}
    \end{minipage}
\end{center}

\paragraph{Cross-Model Validation.}
Additional results in the supplementary material compare GRIMIP across seven foundation models. They show that model scale does not strictly determine tuning performance and that locally deployed smaller models can remain competitive while avoiding API cost and supporting private, offline operation.

\begin{figure}[t]
    \centering
    \includegraphics[width=0.92\linewidth]{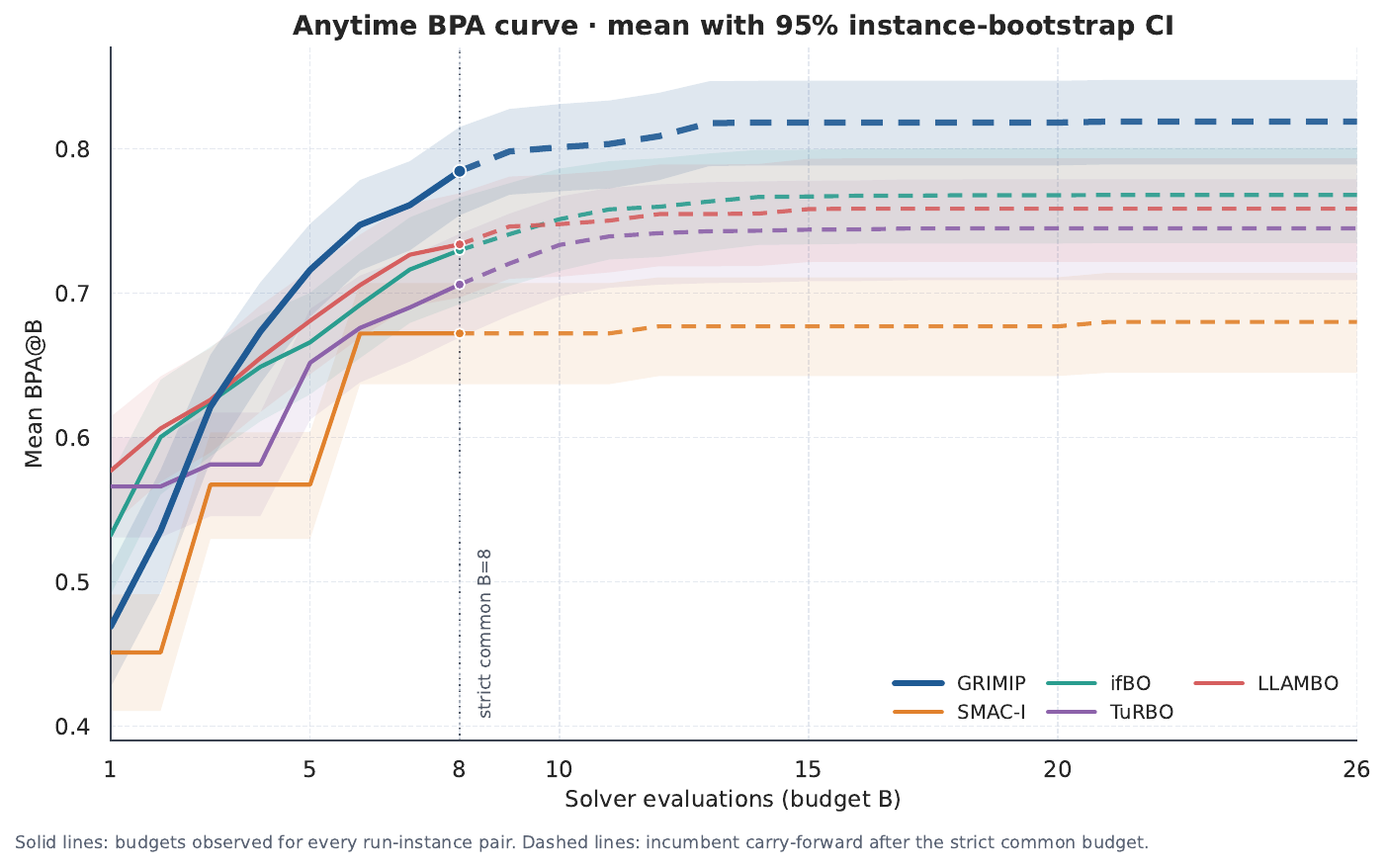}
    \caption{Anytime BPA on the IP dataset. Curves show the mean per-instance BPA and 95\% instance-bootstrap intervals.}
    \label{fig:bpa_anytime_main}
\end{figure}

\paragraph{Convergence Analysis.}
Budgeted PDI Attainment (BPA) complements endpoint PDI by evaluating performance under a finite tuning budget. On IP, GRIMIP achieves the highest mean attainment at the strict common budget $B=8$. Its five-configuration initialization population provides a strong early gain, and the subsequent sequential decisions continue to improve attainment. Figure~\ref{fig:bpa_anytime_main} shows this progression; the BPA definition, statistical protocol, fixed-budget table, and additional attainment profiles are provided in the supplementary material.
The curve further suggests that a total budget of eight evaluations offers a favorable cost--attainment trade-off, reaching roughly $95\%$ of the plateau BPA while avoiding the diminishing returns of additional evaluations.

\paragraph{Search Behavior.}
GRIMIP rapidly concentrates evaluations around promising configurations: its informed initialization identifies a high-value region, and subsequent BO iterations refine it further. This behavior explains its strong performance under short tuning budgets. However, this exploitation-oriented tendency may be detrimental to long-horizon exploration, as the search can concentrate too early and leave promising regions insufficiently examined. The supplementary material provides the complete search-behavior analysis and protocol.

\paragraph{Prediction Calibration and Candidate Ranking.}
Although the LLM's absolute performance predictions are poorly calibrated and its prediction intervals have low empirical coverage, its within-instance candidate ranking remains demonstrably effective and therefore provides a useful ordinal signal for search. Online calibration improves empirical coverage but does not improve final optimization performance under the limited tuning budget; our analysis indicates that calibration overweights uncertainty and consequently spends too much of the budget on exploration. This suggests that GRIMIP benefits more from relative candidate ordering than from calibrated absolute uncertainty. Detailed calibration and candidate-level ranking analyses are provided in the supplementary material under \emph{LLM Ranking Remains Valuable Despite Poor Calibration}.

\section{Conclusion}
This paper presents GRIMIP, marking the pioneering introduction of the LLM4BO paradigm to the task of solver configuration. In essence, our framework can be conceptualized as a hybrid tuning system that fuses \textit{the systematic search of BO} with \textit{the reasoning of a domain expert}. By granting the LLM sufficient autonomy, we effectively overcome the cold-start problem inherent in traditional BO and lower the high barrier typically required for expert-level tuning. Extensive experiments demonstrate that GRIMIP consistently outperforms other strong BO baselines on challenging datasets, achieving superior results with fewer iterations and lower computational costs. Our analysis suggests that this rapid convergence stems from high-quality initialization followed by aggressive exploitation under short budgets. Furthermore, with the support of locally deployed LLMs, we believe this method possesses the potential to replace a wide range of BO-based solver tuning methods in the future. More broadly, the strong priors and high search efficiency of LLMs may be valuable for a wider class of exploration-exploitation tasks, opening a promising direction for future research.

\clearpage
\appendix
\setcounter{secnumdepth}{2}
\section{Experiment Environment} \label{envir}

All computational experiments were performed on a high-performance server running a Linux-based operating system (Ubuntu 20.04.6 LTS). The system was powered by an AMD EPYC 9754 processor, featuring 128 cores capable of handling 256 simultaneous threads, with a standard clock speed of approximately 3.1 GHz. This CPU is equipped with 256 MiB of L3 cache. The server was configured with 251 GiB of system RAM. To ensure fair and reproducible comparisons, all evaluation tasks were strictly limited to a single execution thread. For tasks requiring hardware acceleration, we used a single GPU-class accelerator with 24 GiB of dedicated memory and a vendor-provided software stack. The primary software used for the experiments included Python version 3.8.20 and Gurobi Optimizer version 11.0.1.

\section{Cross-Model Validation}

\begin{figure}[h]
    \centering
    \includegraphics[width=\linewidth]{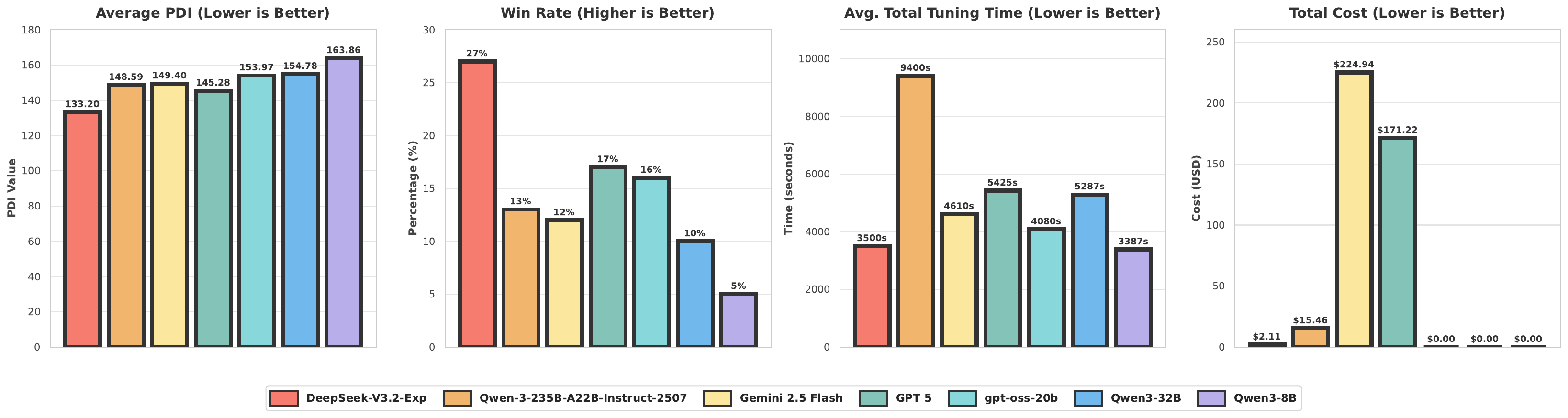}
    \caption{Performance Comparison of LLMs. GRIMIP is benchmarked across seven models, including both API-based services and locally deployed open-source models (gpt-oss-20b, Qwen3-32B, and Qwen3-8B, marked with \$0 cost). Smaller models demonstrate competitive tuning efficiency.}
    \label{fig:llm_comparison_supp}
\end{figure}

To assess generalization, we deployed GRIMIP across seven diverse foundation models. Figure~\ref{fig:llm_comparison_supp} shows that model scale does not strictly determine tuning performance. Locally deployed smaller models enable unlimited iterations without API costs, preserve privacy for sensitive simulation parameters, and support offline operation in restricted environments. When integrated with GRIMIP's structured reasoning framework, these models can compensate for reduced scale through domain-specific guidance.

\section{Convergence Trajectories}

\begin{figure*}[t]
    \centering
    \begin{subfigure}{0.24\textwidth}
        \includegraphics[width=\linewidth]{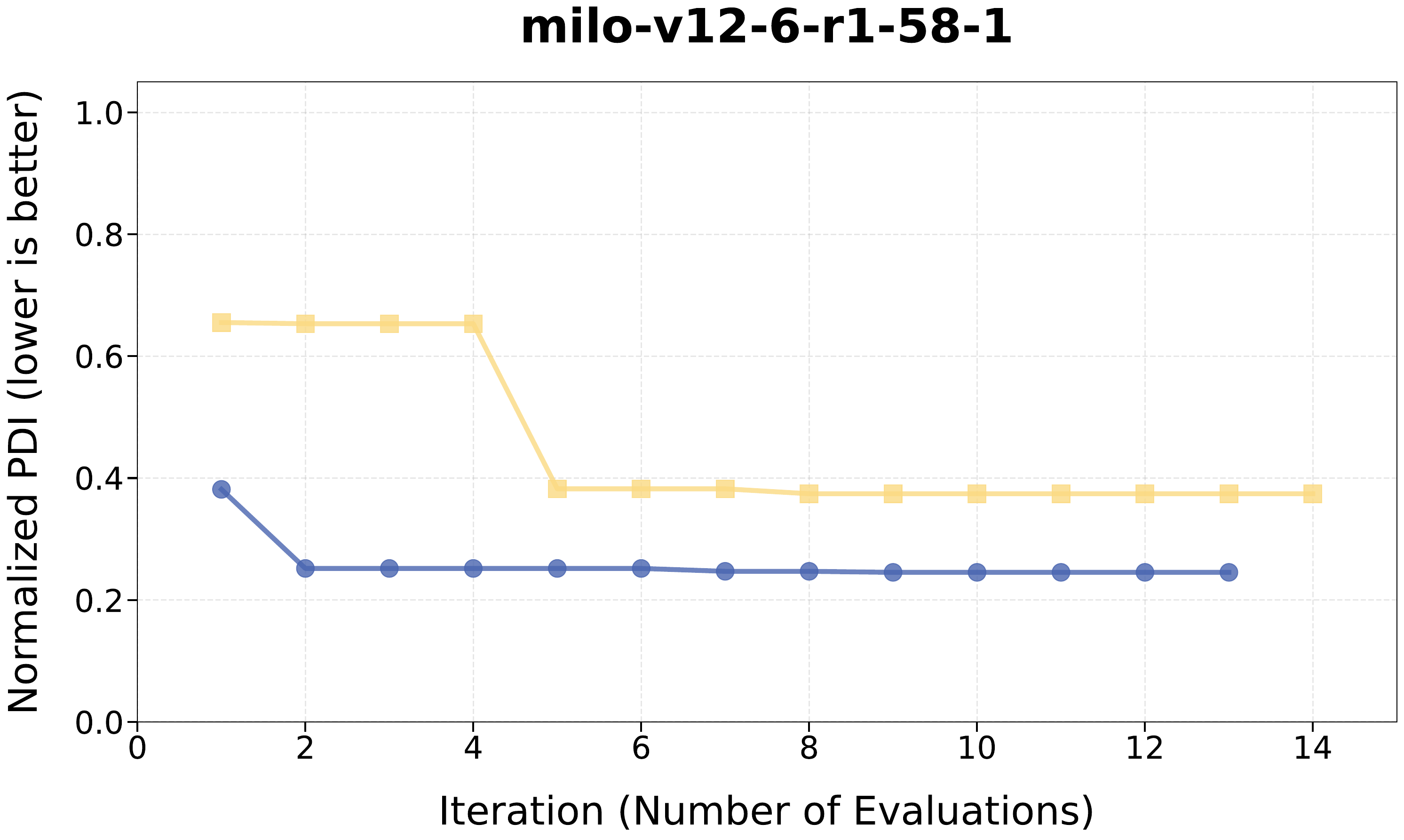}
    \end{subfigure}
    \hfill
    \begin{subfigure}{0.24\textwidth}
        \includegraphics[width=\linewidth]{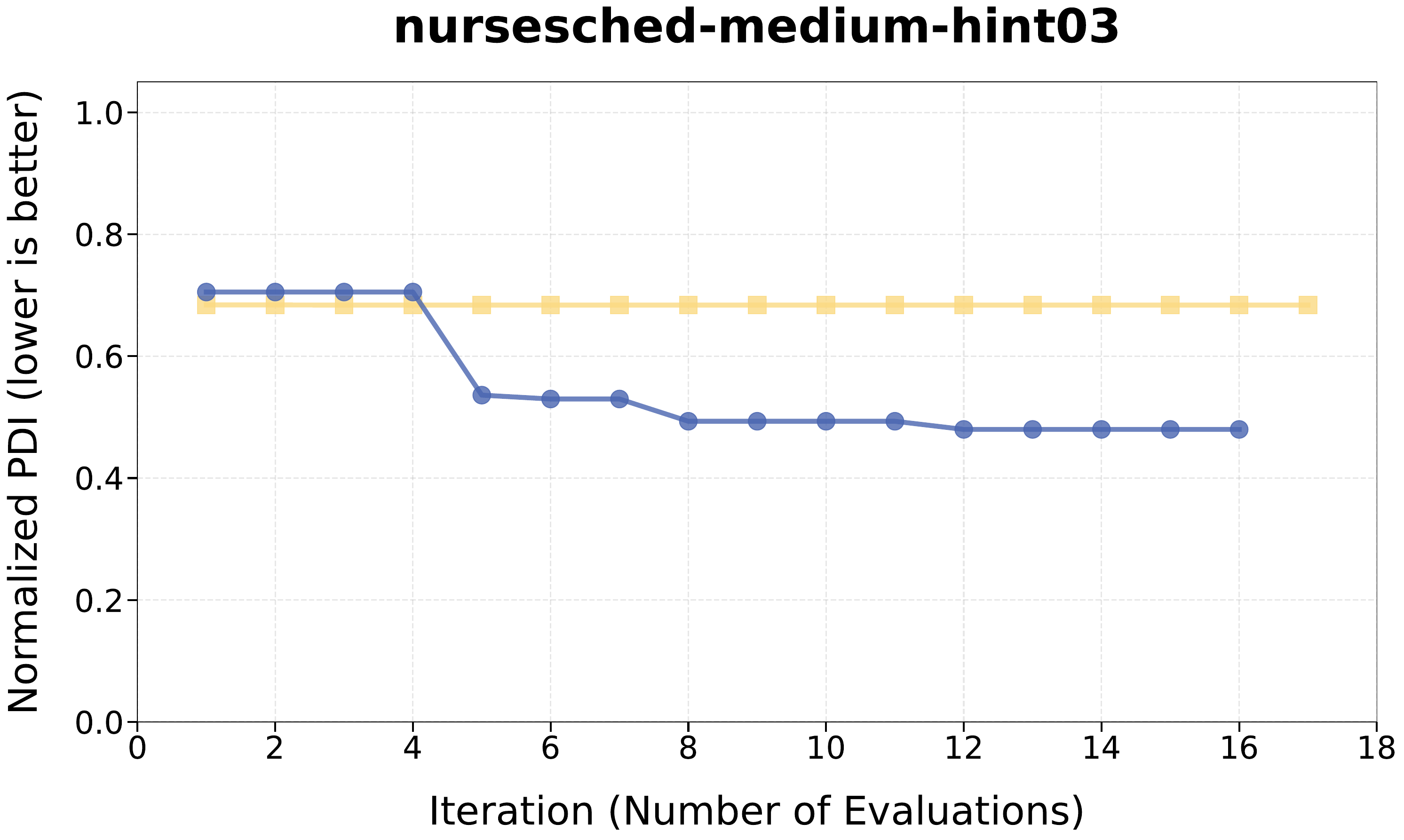}
    \end{subfigure}
    \hfill
    \begin{subfigure}{0.24\textwidth}
        \includegraphics[width=\linewidth]{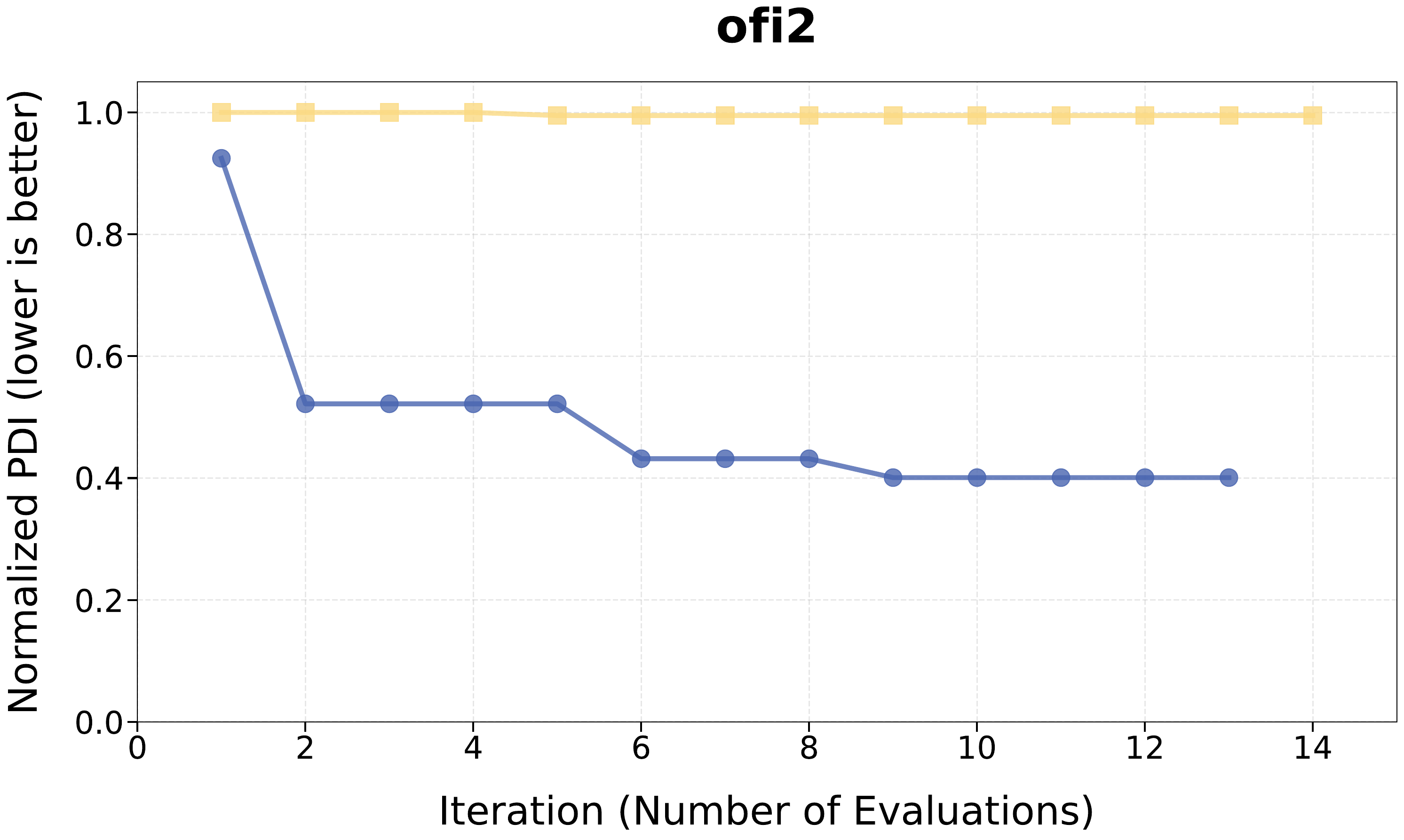}
    \end{subfigure}
    \hfill
    \begin{subfigure}{0.24\textwidth}
        \includegraphics[width=\linewidth]{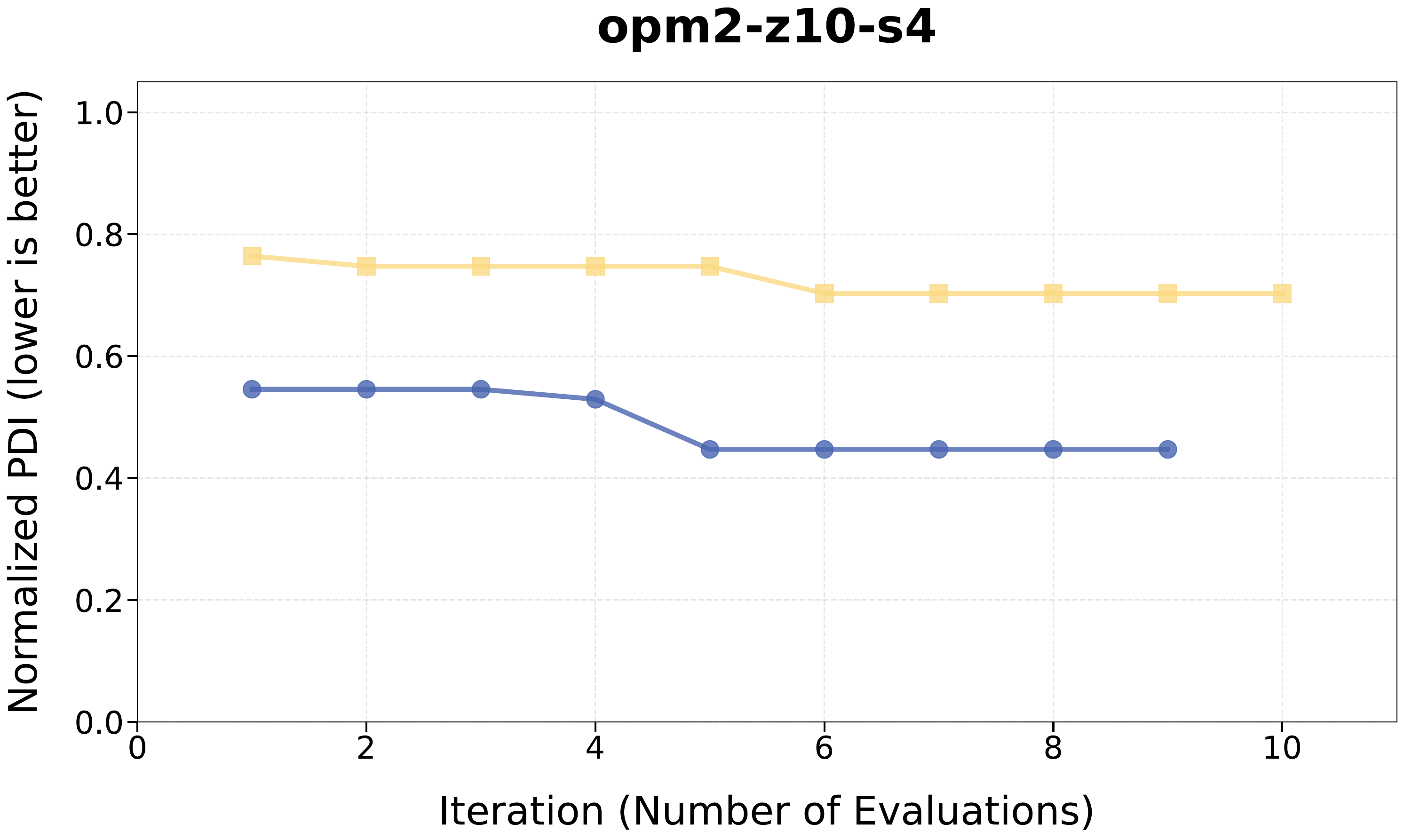}
    \end{subfigure}

    \begin{subfigure}{0.24\textwidth}
        \includegraphics[width=\linewidth]{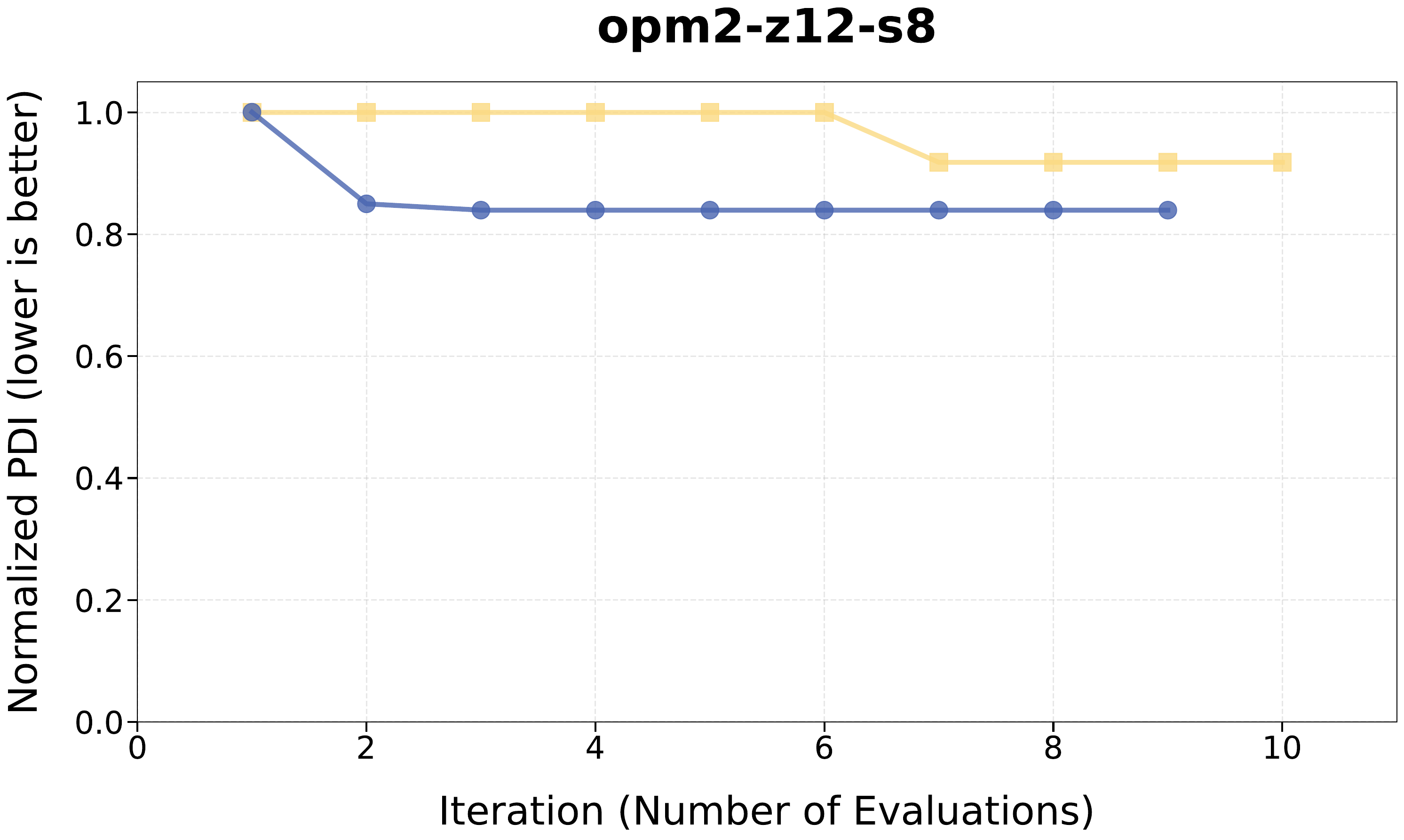}
    \end{subfigure}
    \hfill
    \begin{subfigure}{0.24\textwidth}
        \includegraphics[width=\linewidth]{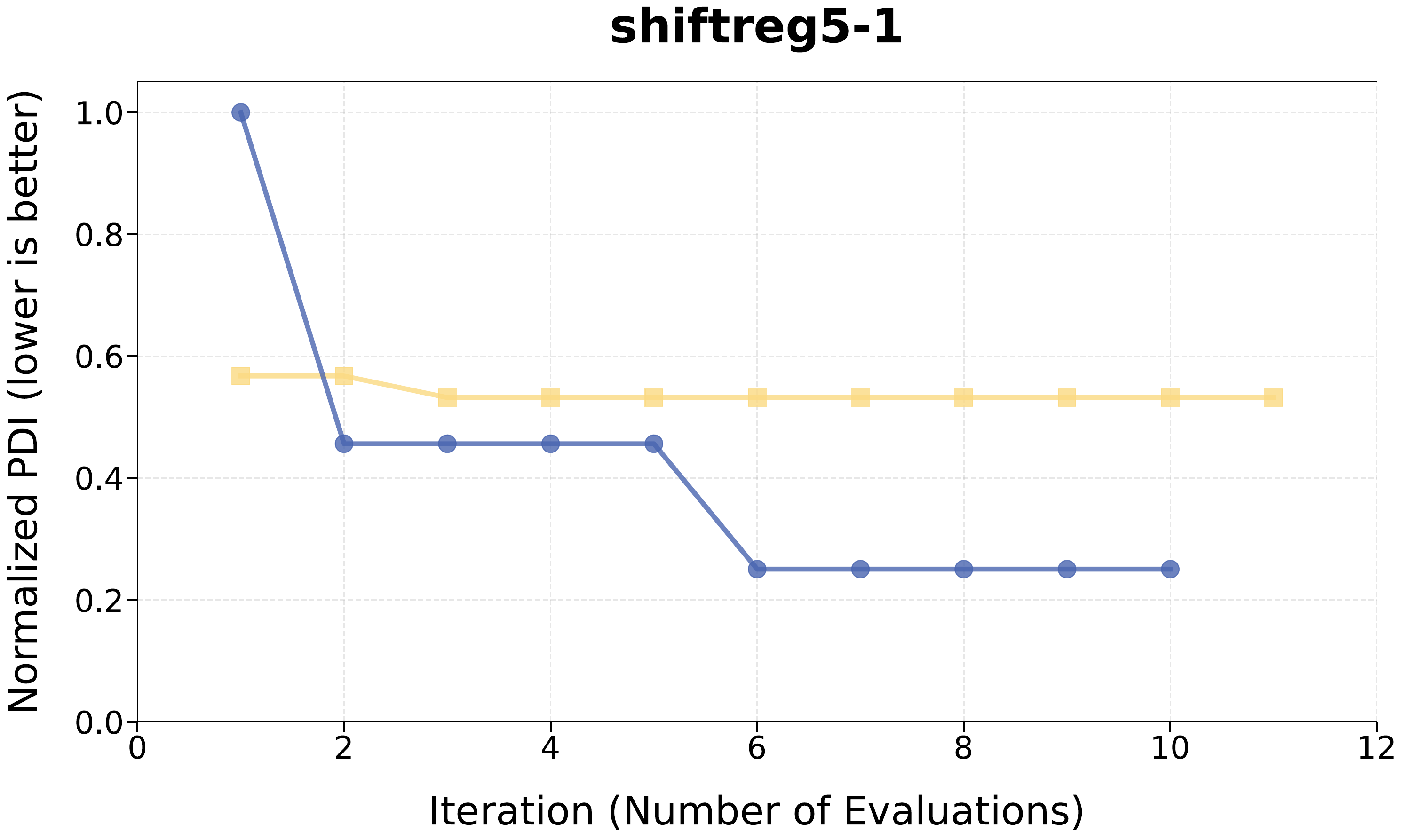}
    \end{subfigure}
    \hfill
    \begin{subfigure}{0.24\textwidth}
        \includegraphics[width=\linewidth]{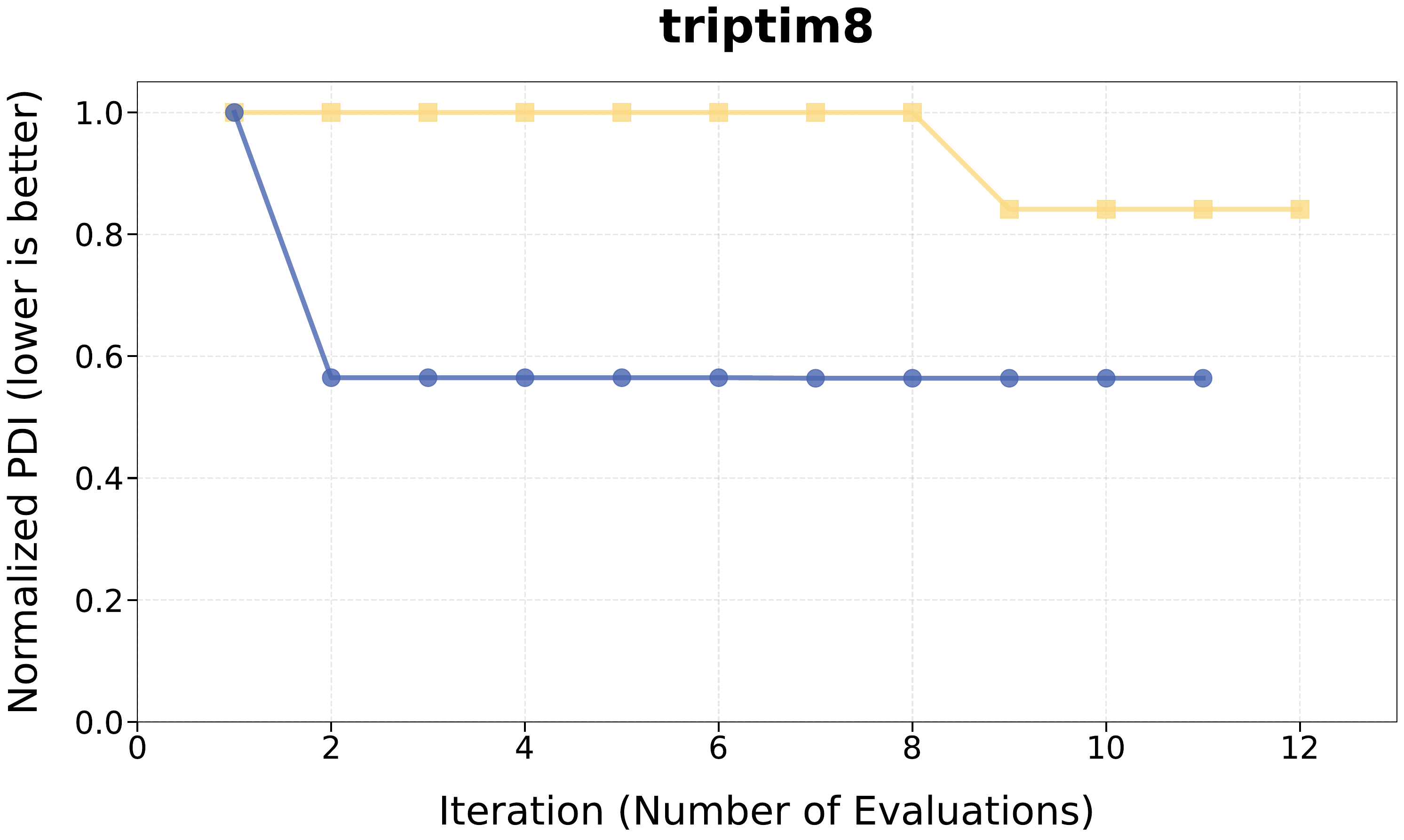}
    \end{subfigure}
    \hfill
    \begin{subfigure}{0.24\textwidth}
        \includegraphics[width=\linewidth]{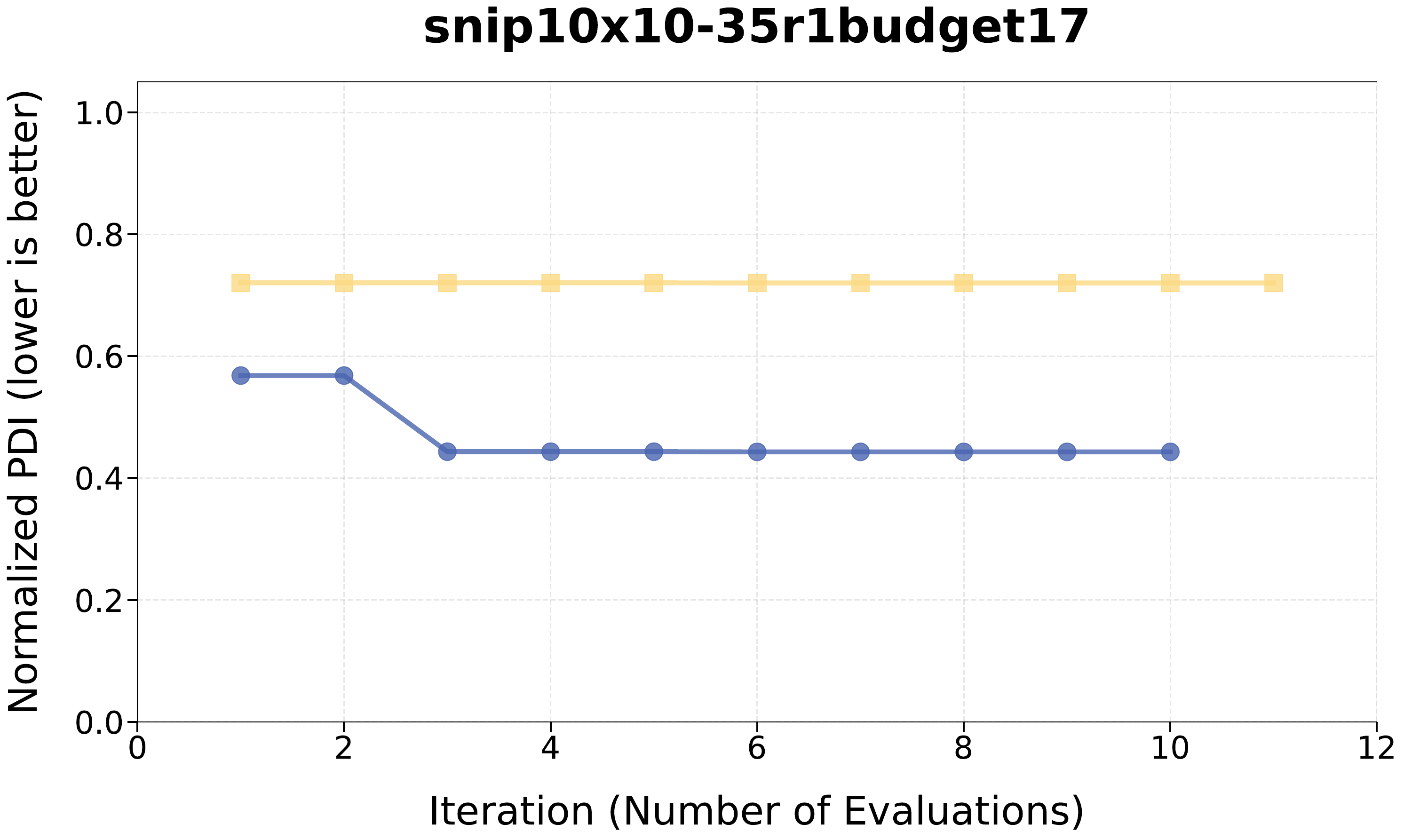}
    \end{subfigure}

    \caption{Convergence trajectories of GRIMIP versus SMAC-I on selected MIPLIB instances. The plots compare SMAC-I (\coloredline{smac_color}) and GRIMIP (\coloredline{grimip_color}). Normalized PDI is shown against the number of function evaluations; lower is better.}
    \label{fig:convergence_grid_supp}
\end{figure*}

\section{LLM Ranking Remains Valuable Despite Poor Calibration}
\label{app:ranking_value}

The LLM's absolute performance predictions and uncertainty estimates are not well calibrated.
The following analyses nevertheless show that its within-instance ranking signal is useful and
becomes stronger as online history accumulates.

\subsection{Complete Candidate-Level Ranking Audit}
\label{app:candidate_audit}

\paragraph{Protocol and labels.}
To audit LLM prediction quality, we reran GRIMIP on all 100 instances of the IP benchmark with
complete candidate logging.  Every online decision group contains the three configurations
proposed by the LLM, their predicted mean ($\mu$) and uncertainty ($\sigma$), and a true PDI
label.  The normally selected candidate inherits its PDI from the optimization trajectory,
while rejected candidates are evaluated separately.  This produced 673 decision groups
(2,019 candidates), including 1,346 newly evaluated rejected candidates.

\begin{table*}[t]
\centering
\small
\caption{Candidate ranking and calibration results.  Pairwise accuracy is computed within each
three-candidate group, so it is not confounded by instance-to-instance PDI scale.}
\label{tab:candidate_ranking}
\begin{tabular}{lrl}
\toprule
\textbf{Metric} & \textbf{Result} & \textbf{Interpretation} \\
\midrule
$\mu$ pairwise accuracy & 70.20\% [67.78, 72.54] & Stable ordinal signal \\
$\mu$ Spearman correlation & 0.433 [0.383, 0.479] & Positive within-group association \\
$\mu$ top-1 accuracy (unique-best groups) & 65.23\% & Useful top-ranked-candidate signal \\
\midrule
Mean predicted $\mu$ / true PDI & 181.07 / 195.34 & Systematic underestimation \\
MAE / RMSE & 28.63 / 40.27 & Poor absolute calibration \\
$\mu\mathbin{\pm}\sigma$ coverage & 48.34\% & Below the nominal 68.27\% \\
\bottomrule
\end{tabular}
\end{table*}

\paragraph{Why poor calibration does not invalidate ranking.}
Calibration asks whether $\mu$ is numerically close to PDI and whether $\sigma$ has the
claimed probabilistic scale.  Ranking asks whether, for two candidates of the same instance,
the better candidate tends to receive the better prediction.  The audit shows that the former
is poor (MAE 28.63 and one-sigma coverage 48.34\%), while the latter remains demonstrably
useful (70.20\% pairwise accuracy).  The distinction is therefore empirical rather than semantic:
the LLM has a useful ordinal signal even though its absolute predictions and uncertainty scale
are not calibrated.

\subsection{Does Candidate Prediction Improve with Online Iterations?}
\label{app:iteration_accuracy}

\paragraph{Analysis design.}
The first candidate prediction after the five warm-start evaluations is defined as online round
1.  We evaluate two notions of improvement: (i) numerical prediction error (MAE, relative MAE,
and one-sigma coverage), and (ii) the ranking induced by $\mu$ (pairwise accuracy and Spearman
correlation).  Because
patience-based early stopping leaves fewer, harder instances in later rounds, we use instance
fixed-effect trends with cluster-robust standard errors, a matched round 1--2 versus 3--4
comparison over all 100 instances, and a matched round 1--4 versus 5--8 comparison over the 31
instances that reached round 8.

\begin{table}[!htbp]
\centering
\scriptsize
\setlength{\tabcolsep}{2pt}
\caption{Instance fixed-effect trend in candidate prediction quality.  Slopes are changes per
online round; confidence intervals are cluster-robust 95\% intervals.  The final column reports
Holm-adjusted p-values.}
\label{tab:iteration_fixed_effects}
\begin{tabular}{lrrr}
\toprule
\textbf{Metric} & \textbf{Slope / round} & \textbf{95\% CI} & \textbf{Holm $p$} \\
\midrule
Group MAE & $-0.799$ PDI & $[-1.213,-0.385]$ & 0.0014 \\
Relative MAE & $-0.510$ pp & $[-0.750,-0.270]$ & 0.0004 \\
One-sigma coverage & $+1.645$ pp & $[+0.725,+2.565]$ & 0.0030 \\
$\mu$ pairwise accuracy & $+1.618$ pp & $[+0.458,+2.777]$ & 0.0202 \\
$\mu$ Spearman correlation & $+0.0338$ & $[+0.0115,+0.0561]$ & 0.0132 \\
\bottomrule
\end{tabular}
\end{table}

The fixed-effect results show a consistent improvement in the mean prediction: each additional
online round reduces MAE by 0.799 PDI and increases $\mu$ pairwise accuracy by 1.618 percentage
points.  One-sigma coverage also rises, but remains below the nominal 68.27\% level. Online
history improves the LLM's mean prediction and its within-instance ordinal ranking.

\begin{table}[!htbp]
\centering
\scriptsize
\setlength{\tabcolsep}{1.5pt}
\caption{Matched-window comparisons.  The first comparison uses all 100 instances; the second
uses the same 31 instances that survived to online round 8.}
\label{tab:iteration_matched}
\begin{tabular}{lrrrrr}
\toprule
\textbf{Metric} &
\shortstack{\textbf{Rounds}\\\textbf{1--2}} &
\shortstack{\textbf{Rounds}\\\textbf{3--4}} &
\shortstack{\textbf{Rounds}\\\textbf{1--4}} &
\shortstack{\textbf{Rounds}\\\textbf{5--8}} &
\shortstack{\textbf{Late--}\\\textbf{early}} \\
\midrule
MAE & 31.51 & 30.43 & 31.98 & 26.29 & $-5.69^{*}$ \\
Relative MAE & 18.50\% & 17.42\% & 18.65\% & 15.20\% & $-3.45$ pp$^{*}$ \\
One-sigma coverage & -- & -- & 39.78\% & 52.42\% & $+12.63$ pp$^{*}$ \\
$\mu$ pairwise & 63.38\% & 68.27\% & 66.80\% & 76.34\% & $+9.54$ pp$^{*}$ \\
$\mu$ Spearman & 0.283 & 0.391 & 0.357 & 0.564 & $+0.207^{*}$ \\
\bottomrule
\end{tabular}

\vspace{1mm}
\raggedright\footnotesize
The asterisk denotes a Wilcoxon or paired-bootstrap test with $p<0.01$ for the round 1--4
versus 5--8 comparison.  In the all-instance round 1--2 versus 3--4 comparison, the direction
is favorable but the $\mu$ changes are not significant ($p=0.117$ for pairwise accuracy and
$p=0.068$ for Spearman).
\end{table}

\FloatBarrier

\section{More Related Work} \label{relatedwork}

\paragraph{Advanced Surrogate Models for Hyperparameter Optimization}
While standard Bayesian Optimization relies on Gaussian Processes (GPs), recent research has sought to enhance expressiveness for complex landscapes. Techniques include Deep Kernel GPs \citep{wilson2016deep, hebbal2018efficient} and Manifold GPs \citep{calandra2016manifold} to better capture non-stationary behaviors. Beyond GPs, Tree-structured Parzen Estimators (TPE) offer a generative approach, modeling $p(h \mid s)$ and $p(h)$ directly, which often scales better to high-dimensional spaces \citep{bergstra2011algorithms, akiba2019optuna}. A more recent trend involves leveraging Transformer architectures as pre-trained priors for BO, such as Prior-Data Fitted Networks (PFNs), which allow for in-context learning of the surrogate surface \citep{chen2022towards, muller2023pfns4bo}. These advanced surrogates aim to refine the trade-off between exploring uncertain regions and exploiting known high-performing parameters.

\paragraph{HPO under Complex Constraints and Objectives}
Standard HPO often assumes a single objective with uniform evaluation costs, but practical solver tuning is more nuanced. To address finite computational budgets, multi-fidelity approximations like Hyperband \citep{li2017hyperband} and FABOLAS \citep{klein2017fast} allocate resources dynamically, discarding poor configurations early. For problems requiring simultaneous optimization of conflicting metrics (e.g., solution quality vs. runtime), multi-objective optimization techniques have been integrated into BO frameworks \citep{daulton2022multi}. Additionally, Multi-task learning \citep{swersky2013multi} allows information transfer across related tuning tasks. In the context of MIP specifically, \citet{sorrell2017tuning} applied statistical experimental design to isolate influential parameters for specific problem classes, though such methods typically lack the flexibility to adapt to new instances without retraining.

\paragraph{LLM-Driven Mechanisms for Optimization}
While the Introduction outlines the general use of LLMs in optimization, recent works have proposed specific mechanisms to ground LLM reasoning. For instance, \citet{zhang2023using} utilize LLMs in a conversational format, iteratively refining suggestions based on feedback history. To tackle the specific challenges of combinatorial optimization, the Hercules framework introduces specialized prompting strategies: ``Core Abstraction Prompting'' to distill knowledge from elite heuristics, and ``Performance Prediction Prompting'' to estimate solution quality before expensive evaluations \citep{wu2025efficient}. Other approaches focus on the surrogate role, such as using LLMs to model the probability of improvement in molecular optimization tasks \citep{ramos2023bayesian}. These methods highlight the potential of structured prompting and hybrid architectures in enhancing the reliability of LLM-based optimizers.

\paragraph{Traditional Heuristic Methods.} Standard automated algorithm configuration methods include fractional factorial design, local search strategies, genetic algorithms, and racing procedures. CALIBRA combines Taguchi's fractional experimental designs with local search to find optimal values for up to five search parameters~\citep{adenso2006fine}. ParamILS employs iterated local search in the parameter configuration space with adaptive capping techniques to accelerate runtime optimization~\citep{hutter2009paramils}. Golden Parameter Search (GPS) exploits the unimodal structure of configuration landscapes to optimize parameters semi-independently in parallel~\citep{pushak2020golden}. GGA introduces a gender-based selection mechanism with AND-OR tree representations to capture parameter dependencies~\citep{ansotegui2009gender}. GGA++ extends this by incorporating random forest surrogate models to predict high-performance parameter regions~\citep{ansotegui2015model}. PyDGGA further distributes GGA across computing clusters using an event-driven architecture~\citep{ansotegui2021pydgga}. The irace package implements iterated racing procedures that use statistical tests to progressively eliminate underperforming configurations~\citep{lopez-ibanez2016irace}.

\section{Implementation and Reproducibility Details}
\label{app:implementation_details}

\subsection{Search Budget and Stopping Rule}

Table~\ref{tab:grimip_settings} gives the complete settings used in the reported GRIMIP experiments. The maximum iteration count denotes the total number of expensive solver evaluations, including the warm-start evaluations. At each adaptive round, the LLM generates three candidates in parallel, predicts a mean and standard deviation for each, and only the candidate with the largest EI is evaluated by Gurobi.

\begin{table*}[t]
\centering
\small
\caption{GRIMIP search settings used in the main experiments.}
\label{tab:grimip_settings}
\begin{tabularx}{\textwidth}{@{}lX@{}}
\toprule
\textbf{Setting} & \textbf{Value} \\
\midrule
LLM & DeepSeek-V3.2-Exp; temperature 0.1; top-$p$ 0.8; top-$k$ 40; maximum output 8,000 tokens \\
ASS & Select at most 6 parameters once before warm-starting; parameter ranges remain fixed; dynamic ASS update disabled \\
Preliminary feature run & 30 seconds with Gurobi defaults; not counted as a configuration evaluation \\
Initial population & $N_{\mathrm{init}}=5$ LLM-generated configurations; all five are evaluated \\
Candidate batch & $K=3$ candidates per adaptive round, using exploitation, exploration, and balanced prompts; one candidate is selected by EI and evaluated \\
Maximum evaluations & 50 total solver evaluations per instance, comprising 5 initial and at most 45 adaptive evaluations \\
Stopping rule & Stop when either 50 total evaluations are reached or 4 consecutive adaptive evaluations fail to strictly improve the incumbent; a strict improvement resets the counter. Failure to obtain a valid candidate also terminates the run. \\
Solver evaluation & Single-threaded Gurobi; 300-second limit per PDI evaluation \\
\bottomrule
\end{tabularx}
\end{table*}

\subsection{ASS Parameter Pool and Fixed-Space Ablation}

ASS selects its instance-specific subset from the 15 parameters in Table~\ref{tab:ass_parameter_pool}. It selects parameter names only; the listed types and ranges are fixed by the implementation.

\begin{table*}[t]
\centering
\small
\caption{Complete Gurobi parameter pool exposed to ASS.}
\label{tab:ass_parameter_pool}
\setlength{\tabcolsep}{5pt}
\begin{tabular}{lll@{\hspace{18pt}}lll}
\toprule
\textbf{Parameter} & \textbf{Type} & \textbf{Range} &
\textbf{Parameter} & \textbf{Type} & \textbf{Range} \\
\midrule
\texttt{MIPFocus}     & int   & $[0,3]$    & \texttt{MIRCuts}       & int & $[-1,2]$ \\
\texttt{Heuristics}   & float & $[0,1]$    & \texttt{CliqueCuts}    & int & $[-1,2]$ \\
\texttt{Cuts}         & int   & $[-1,3]$   & \texttt{ZeroHalfCuts}  & int & $[-1,2]$ \\
\texttt{Presolve}     & int   & $[-1,2]$   & \texttt{BranchDir}     & int & $[-1,1]$ \\
\texttt{Method}       & int   & $[-1,2]$   & \texttt{Symmetry}      & int & $[-1,2]$ \\
\texttt{VarBranch}    & int   & $[-1,3]$   & \texttt{NodeMethod}    & int & $[0,2]$ \\
\texttt{GomoryPasses} & int   & $[-1,100]$ & \texttt{CoverCuts}     & int & $[-1,2]$ \\
\texttt{FlowCoverCuts}& int   & $[-1,2]$   &                       &     &          \\
\bottomrule
\end{tabular}
\end{table*}

The \emph{GRIMIP w/o ASS (16 params)} ablation uses the fixed set
\{\texttt{MIPFocus}, \texttt{Heuristics}, \texttt{Cuts}, \texttt{Presolve},
\texttt{Method}, \texttt{VarBranch}, \texttt{GomoryPasses}, \texttt{CoverCuts},
\texttt{FlowCoverCuts}, \texttt{MIRCuts}, \texttt{CliqueCuts},
\texttt{ZeroHalfCuts}, \texttt{BranchDir}, \texttt{Symmetry},
\texttt{NodeMethod}, \texttt{NodefileStart}\}.
The first 15 parameters use the ranges in Table~\ref{tab:ass_parameter_pool};
\texttt{NodefileStart} is continuous on $[0.1,2.0]$.

\FloatBarrier
\onecolumn
\makeatletter
\global\@colht\textheight
\global\@colroom\textheight
\global\vsize\textheight
\makeatother
\raggedbottom
\subsection{Verbatim Prompt Templates}
\label{app:complete_prompts}

The following listings reproduce the prompt text used by the implementation. Tokens enclosed in double braces denote runtime substitutions, such as the selected parameter space, instance features, optimization history, or a candidate configuration; they are replaced without changing the surrounding text. Markdown markers are part of the prompts.

\promptheading{Automated Space Selection (ASS).}

\begin{lstlisting}[style=prompt]
You are an expert in Gurobi solver optimization with deep understanding of parameter interactions and their impact on different problem types.

## Task: Automated Space Selection (ASS)
Based on the problem instance characteristics, select the most impactful Gurobi parameters to tune and define their search space.

## Available Gurobi Parameters:

## Core MIP Parameters:
- MIPFocus: integer in {0,1,2,3} - MIP solution strategy (0=balance, 1=feasibility, 2=optimality, 3=bound)
- Heuristics: float in [0.0, 1.0] - Fraction of time spent on MIP heuristics
- Cuts: integer in {-1,0,1,2,3} - Global cut aggressiveness (-1=auto, 0=off, 1=conservative, 2=moderate, 3=aggressive)
- Presolve: integer in {-1,0,1,2} - Presolve level (-1=auto, 0=off, 1=conservative, 2=aggressive)
- Method: integer in {-1,0,1,2} - Algorithm for LP/root relaxation (-1=auto, 0=primal simplex, 1=dual simplex, 2=barrier)
- VarBranch: integer in {-1,0,1,2,3} - Variable branching (-1=auto, 0=pseudo reduced-cost, 1=pseudo shadow price, 2=max infeasibility, 3=strong branching)

## Additional Cutting Plane Parameters:
- GomoryPasses: integer in [-1, 100] - Maximum Gomory cut passes (-1=auto, 0=off)
- CoverCuts: integer in {-1,0,1,2} - Cover cut generation (-1=auto, 0=off, 1=conservative, 2=aggressive)
- FlowCoverCuts: integer in {-1,0,1,2} - Flow cover cut generation (-1=auto, 0=off, 1=conservative, 2=aggressive)
- MIRCuts: integer in {-1,0,1,2} - MIR cut generation (-1=auto, 0=off, 1=conservative, 2=aggressive)
- CliqueCuts: integer in {-1,0,1,2} - Clique cut generation (-1=auto, 0=off, 1=conservative, 2=aggressive)
- ZeroHalfCuts: integer in {-1,0,1,2} - Zero-half cut generation (-1=auto, 0=off, 1=conservative, 2=aggressive)

## Search Strategy Parameters:
- BranchDir: integer in {-1,0,1} - Branch direction preference (-1=down, 0=auto, 1=up)
- Symmetry: integer in {-1,0,1,2} - Symmetry detection/breaking (-1=auto, 0=off, 1=conservative, 2=aggressive)
- NodeMethod: integer in {0,1,2} - Algorithm for non-root nodes (0=primal simplex, 1=dual simplex, 2=barrier)

## Current Instance Features:
{{INSTANCE_FEATURES}}

## Selection Criteria:
1. **Impact Analysis**: Select parameters that will have the most significant impact on this specific instance
2. **Interaction Effects**: Consider parameter synergies and conflicts
3. **Instance-Specific**: Tailor selection based on problem characteristics:
   - Problem size (variables, constraints, non-zeros)
   - Problem type (integer ratio, binary ratio)
   - Problem structure (constraint matrix density, tree depth)
   - Initial solve characteristics (root gap, node processing rate)

## Selection Guidelines:
- For large-scale problems: prioritize memory and algorithmic method parameters
- For high integer ratio: focus on branching, heuristics, and cutting strategies
- For dense constraint matrices: consider presolve and LP method parameters
- For problems with poor root relaxation: emphasize cutting plane parameters
- For quick initial feasibility: prioritize heuristic and MIPFocus parameters

## Constraint: Select at most 6 parameters to tune.

## Important: Parameter Range Policy
- You should ONLY select which parameters to tune
- DO NOT adjust the min/max bounds - all parameters will use their original ranges from the configuration
- The ranges shown in the parameter list above are fixed and cannot be modified
- Focus on selecting the most impactful parameters, not on adjusting their ranges

## Output Format:
Return a JSON object with selected parameters (you only need to list parameter names and reasons):
```json
{
  "selected_params": {
    "ParameterName": {
      "reason": "Brief explanation why this parameter is selected"
    },
    ...
  },
  "selection_rationale": "Overall strategy explanation"
}
```

Note: You do NOT need to specify type/min/max - the system will automatically use the original ranges from configuration.

Please analyze the instance characteristics and select the most suitable parameters for tuning.
\end{lstlisting}

\promptheading{Warm-starting.}

\begin{lstlisting}[style=prompt]
You are a world-class Gurobi parameter optimization expert with deep understanding of MIP solver internals and parameter interactions. Please generate 5 strategically diverse initial parameter configurations for Bayesian optimization.

## Tunable parameter space:
{{SELECTED_PARAMETER_SPACE}}

## Current MIP instance features:
{{INSTANCE_FEATURES}}

## Strategic parameter selection guidelines:

Based on the instance features above, please design 5 configurations that are:

1. **Instance-aware**: Tailor parameters to instance characteristics:
   - For large-scale instances: Method=1 (Dual Simplex) is memory-efficient; Method=2 (Barrier) is often faster but uses more memory
   - For high integer ratio: balance between heuristics and systematic search (MIPFocus, Heuristics)
   - For instances with good root relaxation (small initial gap): may focus on optimality (MIPFocus=2)
   - For hard instances (large gap, few solutions in 30s): try aggressive settings (Cuts=3, Heuristics=0.8)

2. **Parameter synergies**: Consider interactions:
   - Method=2 (Barrier) typically benefits from stronger Presolve and Cuts settings
   - MIPFocus=3 (bound improvement) pairs well with VarBranch=3 (strong branching)
   - MIPFocus=1 (feasibility) pairs well with higher Heuristics values
   - High Cuts values (2-3) may require careful Presolve settings to maintain numerical stability
   - For deep search trees: consider stronger branching (VarBranch=3) or different MIPFocus

3. **Diversity through strategic coverage**:
   - Maximize configuration diversity using distance metrics:
     * For integer parameters: Hamming distance (different categorical choices)
     * For continuous parameters: normalized Euclidean distance
   - Avoid dominated configurations (one strictly worse than another across all dimensions)
   - Ensure each configuration has a distinct "strategy profile":
     * Aggressive exploration (high Heuristics, aggressive Cuts)
     * Bound-focused (MIPFocus=3, strong branching)
     * Feasibility-focused (MIPFocus=1, high Heuristics, conservative Cuts)
     * Balanced/default-like (moderate settings)
     * Problem-specific tuned (based on instance features)

4. **Evidence-based reasoning**:
   - DO NOT uniformly sample - each parameter value should have a strategic rationale
   - Justify your choices based on instance features (even if implicit)
   - Avoid "random" combinations that ignore parameter semantics

## Output requirements:
1. Return a JSON array with exactly 5 configurations
2. Each configuration MUST include ALL {{D}} parameters: {{SELECTED_PARAMETER_NAMES}}
3. Each configuration MUST NOT include any other parameters beyond the {{D}} listed above
4. Ensure maximum diversity (avoid similar or dominated configurations)
5. No explanations - only the JSON array

Example format (you must include all parameters shown in the tunable parameter space above):
```json
[
  {{EXAMPLE_CONFIGURATION_AT_LOWER_BOUNDS}},
  ...
]
```

CRITICAL CONSTRAINTS:
- Each configuration object must contain EXACTLY these {{D}} parameters: {{SELECTED_PARAMETER_NAMES}}
- DO NOT add any other Gurobi parameters (e.g., PreDual, AggFill, etc.) - only use the parameters listed above
- DO NOT invent new parameter names - strictly use only the {{D}} parameters from the tunable parameter space

Please return the JSON array directly without any other text.
\end{lstlisting}

\promptheading{Candidate generation.}
One instance of the prompt is issued for each of the three candidates. The \texttt{\{\{STRATEGY\}\}} token is replaced verbatim by the corresponding line below.

\begin{lstlisting}[style=prompt]
Candidate 1:
Focus on **exploitation**: Propose a configuration similar to the best-performing ones, with small refinements to parameter values. Stay close to proven good regions.

Candidate 2:
Focus on **exploration**: Propose a configuration in an under-explored region of the parameter space. Try parameter combinations that are significantly different from history.

Candidate 3:
Use a **balanced strategy**: Combine insights from good configurations with moderate exploration. Try a configuration that bridges between explored and unexplored regions.
\end{lstlisting}

\begin{lstlisting}[style=prompt]
You are a world-class Gurobi performance tuning expert with deep understanding of MIP solver internals. Based on the optimization history, please propose a strategically informed new parameter configuration.

## Tunable parameter space:
{{SELECTED_PARAMETER_SPACE}}

## Optimization history:
{{OPTIMIZATION_HISTORY}}

## Your specific strategy for this candidate (Candidate #{{CANDIDATE_INDEX}}):
{{STRATEGY}}

## Strategic requirements:

1. **Learn from history**:
   - Identify which parameter combinations performed well
   - Understand why they worked based on parameter interactions
   - Avoid repeating poor-performing regions

2. **Parameter synergies** (critical):
   - Method=2 (Barrier) → typically benefits from appropriate Presolve and Cuts settings (fast but memory-intensive)
   - Method=1 (Dual Simplex) → memory-efficient, especially for large-scale instances
   - MIPFocus=3 (bound) → pairs well with VarBranch=3 (strong branching)
   - MIPFocus=1 (feasibility) → pairs well with higher Heuristics (0.6-0.9)
   - Balance Cuts and Presolve settings to maintain numerical stability

3. **Diversity requirement**:
   - This is candidate #{{CANDIDATE_INDEX}} of 3
   - Ensure this candidate is DIFFERENT from other candidates you might generate
   - Vary at least 3-4 parameters significantly to ensure diversity

4. **Target-driven**:
   - Goal: achieve performance score {{TARGET_SCORE}} or better
   - Propose configurations with high expected improvement potential

5. **Avoid dominated solutions**:
   - Don't propose configs that are clearly worse than existing ones
   - Each parameter choice should have strategic justification

Return format (JSON only, no explanations):

CRITICAL CONSTRAINTS:
- You MUST include ALL {{D}} parameters: {{SELECTED_PARAMETER_NAMES}}
- You MUST NOT include any other parameters beyond these {{D}} parameters
- DO NOT add parameters like PreDual, AggFill, or any other Gurobi parameters not listed above
- Only use the exact parameter names from the tunable parameter space

```json
{{JSON_OBJECT_WITH_ALL_SELECTED_PARAMETERS}}
```

Please return only the JSON object with EXACTLY {{D}} parameters listed above, no other text.
\end{lstlisting}

\promptheading{LLM surrogate.}

\begin{lstlisting}[style=prompt]
You are a world-class Gurobi performance tuning expert. Based on the following optimization history, please predict the performance score and uncertainty estimation of the new configuration.

Optimization history:
{{OPTIMIZATION_HISTORY}}

New configuration to predict:
{{CANDIDATE_CONFIGURATION}}

[Analysis ID: {{EIGHT_CHARACTER_CANDIDATE_HASH}}]

Current MIP instance features:
{{INSTANCE_FEATURES}}

Please combine instance features, historical records, and candidate configurations for prediction.

Based on patterns and trends in historical data, please predict the performance score (smaller values are better) and prediction uncertainty of this new configuration.

Considerations:
1. Performance of similar parameter configurations in historical records
2. Interaction effects between parameters
3. Trends observed from the data
4. Prediction confidence and uncertainty
5. Impact of instance features on parameter effects

Please return a JSON object containing the predicted mean and standard deviation (std):

```json
{"mean": predicted_performance_score, "std": uncertainty_estimation_standard_deviation}
```

Note:
- mean should be your best estimate of the performance score
- std should reflect prediction uncertainty, larger values indicate greater uncertainty
- If historical data is limited or configuration differs significantly from history, std should be relatively large
- If there is sufficient similar historical data support, std can be relatively small

Return only the JSON object, no other text.
\end{lstlisting}

\promptheading{Optional dynamic ASS update.}
This prompt belongs to the optional dynamic-ASS mode and was not invoked in the reported main experiments, which keep the six-parameter subset fixed after initial selection. Here \texttt{\{\{AVAILABLE\_GUROBI\_PARAMETERS\}\}} expands to exactly the 15-parameter block printed in the ASS prompt above.

\begin{lstlisting}[style=prompt]
You are an expert in adaptive Gurobi parameter tuning. Based on the optimization history, update the parameter space for more effective search.

## Current Parameter Space (Selected for optimization):
{{CURRENT_PARAMETER_SPACE}}

## All Available Gurobi Parameters:
{{AVAILABLE_GUROBI_PARAMETERS}}

## Optimization History Analysis:
- Total evaluations: {{NUMBER_OF_EVALUATIONS}}
- Best score achieved: {{BEST_SCORE}}
- Average score: {{AVERAGE_SCORE}}
- Improvement trend: {{IMPROVEMENT_TREND}}

## Best Configuration Found:
{{BEST_CONFIGURATION}}

## Parameter Impact Analysis:
{{PARAMETER_IMPACT_ANALYSIS}}

## Update Strategy: ADAPTIVE
- Narrow ranges for parameters showing clear optimal regions
- Expand ranges for under-explored parameters
- Consider removing parameters with minimal impact
- Add new parameters from the "All Available Gurobi Parameters" list if current ones plateau
- Balance exploration and exploitation based on optimization progress

## Instance Characteristics:
{{INSTANCE_FEATURE_SUMMARY}}

## Update Guidelines:
1. **Evidence-based decisions**: Base changes on observed performance patterns
2. **Maintain diversity**: Don't over-narrow too quickly
3. **Consider interactions**: Parameters may work better in combination
4. **Problem-specific**: Tailor to instance characteristics
5. **Incremental adjustments**: Make gradual changes to avoid losing good regions

## Output Format:
Return updated parameter space as JSON:
```json
{
  "selected_params": {
    "ParameterName": {
      "type": "int" or "float",
      "min": new_minimum,
      "max": new_maximum,
      "reason": "Why this update"
    }
  },
  "selection_rationale": "Overall update strategy explanation"
}
```

Analyze the history and provide updated parameter space:
\end{lstlisting}

\twocolumn
\flushbottom

\section{Empirical Search-Behavior Analysis} \label{sec:search_behavior}

We complement the outcome-level evaluation with a trajectory-level audit of how the LLM--EI controller searches the configuration space.  The analysis covers 100 IP trajectories, comprising 1,168 solver evaluations and 668 adaptive proposals after the five-point warm start.  Each instance uses an ASS-selected six-parameter space (seven distinct six-parameter variants occur across the analyzed instances), a 300-second PDI evaluation budget, and the same early-stopping patience of four non-improving adaptive evaluations.  PDI is minimized throughout.  Using these trajectories, we analyze search-space contraction, attraction toward the incumbent, and the controller's response to strong improvements.  The results show a consistent transition from broad initialization to incumbent-centered local search, with stronger localization after meaningful improvements.

This is an analysis of the \emph{selected and evaluated} LLM--EI trajectory, not of raw LLM generations in isolation: the logs retain the configuration selected by EI but not the two candidates rejected at each adaptive round.  Thus, the results characterize the behavior of the complete GRIMIP search controller.

\subsection{Metrics and Protocol}

For two configurations $x$ and $z$ in an instance-specific six-dimensional space, we use a mixed Gower-style distance
\begin{equation}
    d_{\mathrm{mix}}(x,z)=\frac{1}{6}\sum_{j=1}^{6}\delta_j(x_j,z_j).
\end{equation}
For categorical algorithm choices (\texttt{MIPFocus}, \texttt{VarBranch}, and \texttt{NodeMethod}), $\delta_j$ is an indicator of unequal values.  For continuous and ordered parameters, it is the absolute difference normalized by that parameter's allowed range.  As a robustness check, we also use the Hamming distance, i.e., the fraction of the six parameter values that differ.

The initial-search set contains the first five warm-start configurations.  The terminal neighborhood contains the final incumbent and the last four adaptive evaluations.  We define a proposal as \emph{local} when $d_{\mathrm{mix}}\leq 0.10$ and it changes at most two of the six parameters relative to the incumbent immediately before the proposal.  This conservative binary definition is used only for interpretability; all principal conclusions are also supported by the continuous-distance analyses.  We report paired Wilcoxon tests over instances and 20,000 instance-level (or instance-cluster) bootstrap replications for confidence intervals.

\begin{table*}[t]
\centering
\scriptsize
\caption{Summary of search-behavior evidence on the 100 IP trajectories. Distance reductions are paired within instances.  A strong record is a relative PDI improvement of at least $5\%$ over the incumbent immediately before it.}
\label{tab:search_behavior_summary}
\begin{tabularx}{\textwidth}{@{}lXCCX@{}}
\toprule
\textbf{Question} & \textbf{Metric} & \textbf{Reference} & \textbf{Observed} & \textbf{Evidence} \\
\midrule
Initial coverage & Mean mixed dispersion & Initial five: 0.579 & Terminal neighborhood: 0.226 & $61.0\%$ lower; 100/100 contract \\
Robustness & Mean Hamming dispersion & Initial five: 0.948 & Terminal neighborhood: 0.525 & $44.6\%$ lower; 100/100 contract \\
Incumbent attraction & Mean proposal distance & To incumbent: 0.178 & To other history: 0.435 & 82.6\% closer to incumbent \\
Nearest-neighbor test & Incumbent is nearest history point & Geometry-null: 13.8\% & Observed: 53.0\% & $3.8\times$ the null rate \\
Strong-record response & Next proposal is local & No record: 63.2\% & $\geq5\%$ record: 78.6\% & Difference $+15.3$ pp, 95\% CI $[+4.1,+25.5]$ \\
Final-best event & Distance to final-best configuration & Four before: 0.328 & Four after: 0.163 & $50.2\%$ lower; 52/65 contract \\
\bottomrule
\end{tabularx}
\end{table*}

\subsection{From Broad Initialization to Incumbent-Centered Search}

\begin{figure*}[t]
    \centering
    \includegraphics[width=0.96\textwidth]{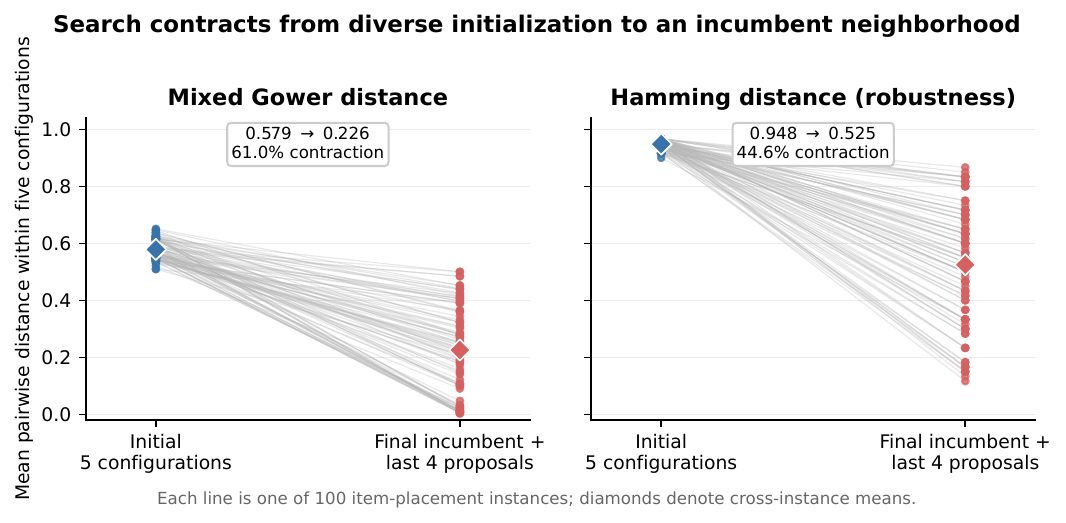}
    \caption{The selected trajectory contracts after diverse initialization. Each line is one IP instance.  The initial set consists of the five LLM warm-start configurations; the terminal set consists of the final incumbent and the last four adaptive evaluations.  The result is consistent under the mixed Gower-style distance and the Hamming-distance robustness check.}
    \label{fig:search_behavior_phase}
\end{figure*}

Figure~\ref{fig:search_behavior_phase} shows a pronounced two-stage trajectory.  The mean mixed dispersion of the initial five configurations is $0.579$, whereas the final-incumbent neighborhood has dispersion $0.226$, a $61.0\%$ reduction.  Every one of the 100 trajectories contracts under this metric (paired Wilcoxon $p=3.90\times10^{-18}$).  The conclusion does not depend on our mixed-distance construction: Hamming dispersion decreases from $0.948$ to $0.525$ ($44.6\%$ reduction; $p=3.89\times10^{-18}$).

The adaptive controller is strongly attracted to the incumbent rather than simply to any previously sampled configuration.  Across the 668 adaptive proposals, the mean distance to the current incumbent is $0.178$ (median $0.025$), compared with $0.435$ to the other historical configurations; 82.6\% of proposals are closer to the incumbent than to the mean of the remaining history.  The current incumbent is the nearest historical configuration for 53.0\% of proposals.  If the incumbent label were geometrically unrelated to the proposal, the corresponding tie-aware probability would be only 13.8\%.  At the same time, the trajectory does not collapse to a purely local search: 405/668 proposals (60.6\%) satisfy the conservative local-proposal definition, leaving 39.4\% as non-local proposals.

\subsection{Strong Improvements Trigger More Local Follow-up}

The controller becomes particularly local after a meaningful improvement.  Excluding the first adaptive round, we compare the next proposal after an adaptive record with the next proposal after a non-record.  Following a record that improves PDI by at least 5\%, the next proposal has mean distance $0.114$ to the incumbent, compared with $0.172$ after a non-record.  More directly, its local-proposal rate rises from 63.2\% to 78.6\%, a $+15.3$ percentage-point difference with an instance-cluster bootstrap 95\% CI of $[+4.1,+25.5]$ percentage points.  The same qualitative pattern persists when the strong-record threshold is varied from 1\% to 20\%.

We observe the same re-centering around the best configuration eventually reached in the run.  In the 65 instances whose final best configuration is found adaptively, the average distance to that configuration declines from $0.328$ over the preceding four proposals to $0.163$ over the following four proposals (52/65 instances; paired Wilcoxon $p=7.66\times10^{-8}$).  Moreover, 47/65 (72.3\%) of these adaptive final-best configurations are themselves local modifications of the preceding incumbent.

\begin{figure*}[t]
    \centering
    \includegraphics[width=0.98\textwidth]{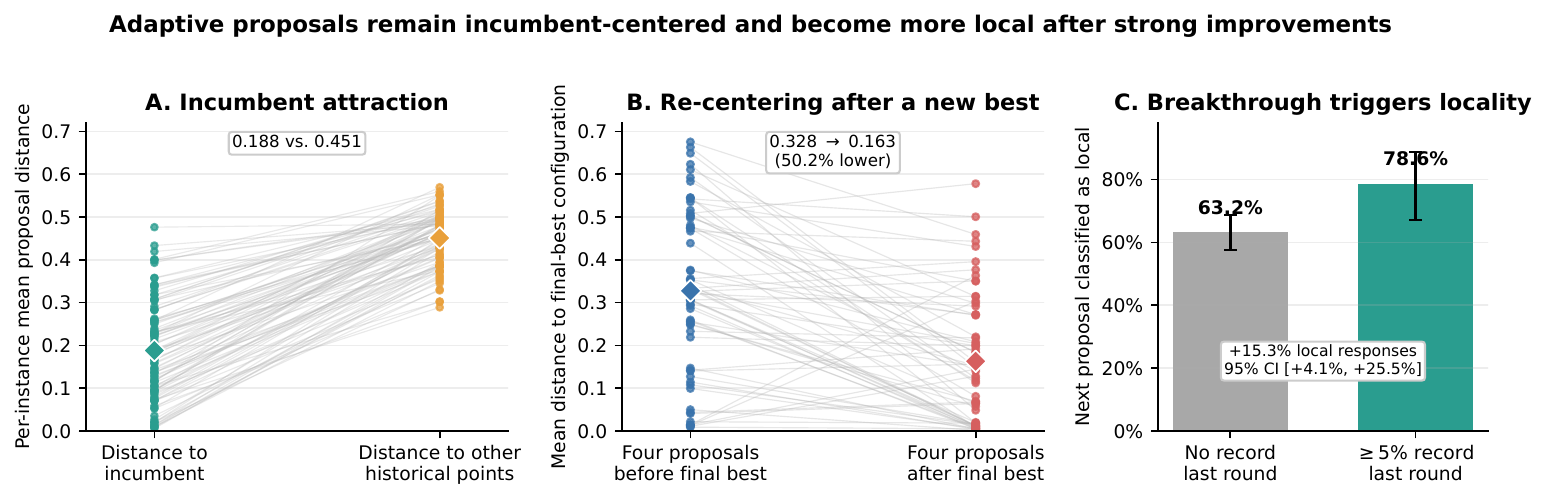}
    \caption{Incumbent attraction and event-triggered locality. (A) Adaptive proposals are substantially closer to the incumbent than to other historical configurations. (B) When the final best is found adaptively, subsequent configurations concentrate around it. (C) A relative PDI record of at least $5\%$ increases the probability that the next proposal is local. Error bars in (C) are 95\% instance-cluster bootstrap intervals.}
    \label{fig:search_behavior_response}
\end{figure*}

\subsection{Which Parameters Are Modified?}

Locality preserves the high-level solution strategy while fine-tuning numerical controls.  Among the 400 configurations evaluated after the final incumbent is found, the exact retention rates are 73.5\% for \texttt{MIPFocus}, 68.3\% for \texttt{VarBranch}, and 72.8\% for \texttt{Cuts}.  In contrast, \texttt{Heuristics} is frequently adjusted, but its median absolute change is only 0.05 on its $[0,1]$ range.  \texttt{GomoryPasses} is also often adjusted, with median absolute change of three passes.  Thus, local exploitation is not mere configuration replay: it generally retains the discrete solver-strategy skeleton while making small numerical refinements.

Importantly, local proposals are more frequent but are not demonstrably more successful per evaluation.  Local and non-local proposals produce a new incumbent at almost identical rates (18.27\% versus 18.25\%); the cluster-bootstrap 95\% CI for their difference is $[-5.17,+4.96]$ percentage points.  This supports an incumbent-centered \emph{mixture} of local refinements and occasional long jumps, rather than a policy that disables exploration once a promising region is identified.

\subsection{Independent Validation on BBOB}
\label{sec:bbob_search_behavior}

To test whether the value of the structured LLM--BO controller extends beyond MIP
parameter tuning, we conducted a controlled study on ten noiseless COCO/BBOB
functions (\(f_1,f_2,f_3,f_5,f_6,f_8,f_{10},f_{12},f_{15},f_{20}\)). We use
instance 1 in dimension \(d=10\), five repetitions per function, and a budget of
60 evaluations (\(6d\)) per run. The comparison includes LLM-direct, LLM-BO,
SMAC 2.4.0, HEBO 0.3.6 with a random-forest surrogate, and uniform random
search. Both LLM variants use the same DeepSeek-V4-Flash backend. LLM-direct
proposes evaluated points directly from the observation history; LLM-BO instead
generates a candidate pool, predicts a mean and uncertainty for each candidate,
and applies EI before evaluation.

We evaluate progress using the official BBOB optimum \(f_{\mathrm{opt}}\), read
only after the searches have finished. For every trajectory, we compute
\(\Delta f(e)=f_{\mathrm{best}}(e)-f_{\mathrm{opt}}\) and record first hitting
times for 51 logarithmically spaced targets in \([10^{-8},10^2]\). The ECDF at
a budget is the fraction of (function, run, target) tuples reached, so functions
with large raw objective scales cannot dominate the aggregate.

\begin{figure*}[t]
    \centering
    \includegraphics[width=0.98\textwidth]{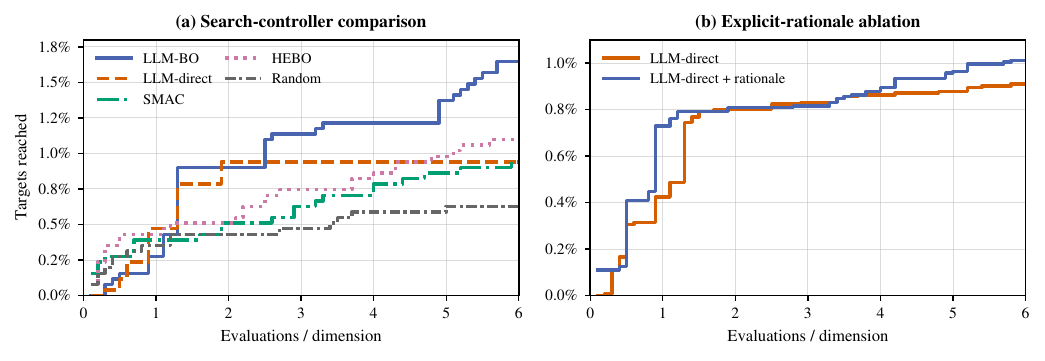}
    \caption{Official-target search behavior on 10-dimensional BBOB.
    (a) Comparison of the structured LLM-BO controller with direct LLM
    proposals and three non-LLM baselines over ten functions, five runs per
    function, and 60 evaluations per run. (b) Paired ablation of requiring an
    explicit mathematical rationale from LLM-direct, with 250 searches per
    condition. Both panels show the fraction of (function, run, target) tuples
    reached over 51 official BBOB targets; higher is better.}
    \label{fig:bbob_search_behavior}
\end{figure*}

At 60 evaluations, LLM-BO reaches \(1.65\%\) of all target tuples, the highest
value among the five methods, compared with \(1.10\%\) for HEBO, \(0.94\%\) for
both LLM-direct and SMAC, and \(0.63\%\) for random search. For the loose target
\(\Delta f\leq 10^2\), the corresponding hit rates are \(24\%\), \(20\%\),
\(18\%\), \(18\%\), and \(14\%\), respectively. LLM-BO also obtains the lowest
final experiment-relative simple regret on six of the ten functions, and its
mean final regret is \(29.1\%\) lower than that of LLM-direct. Because raw
regret scales vary greatly across BBOB functions, we treat the official target
ECDF as the primary aggregate evidence.

We further isolate whether merely asking for more explicit reasoning improves
the direct LLM search. Across five paired trials on the same ten functions
(250 searches per condition), requiring a mathematically grounded rationale
changes the ECDF at 60 evaluations from \(0.91\%\) to \(1.01\%\), a difference
of \(0.10\) percentage points (paired block-bootstrap 95\% CI
\([-0.04,0.27]\) percentage points; block sign-flip \(p=0.2512\)). Although
the mean \(\Delta f\)-AUC improves by \(5.6\%\), the final raw
\(\Delta f\) mean worsens by \(11.5\%\). The rationale condition also produces
\(2.12\times\) as many output characters and incurs nine repair/error batches,
including seven timeouts, versus none without rationales. Thus, explicit
verbosity alone yields no reliable target-attainment gain; the evidence is
instead consistent with the benefit coming from the structured
proposer--surrogate--acquisition loop.

\FloatBarrier
\section{Budgeted PDI Attainment (BPA)} \label{sec:bpa_appendix}

\subsection{Definition and Statistical Protocol} \label{sec:bpa_definition}

The conventional final PDI comparison does not reveal how quickly a tuning
method reaches a good configuration under a finite evaluation budget.  We
therefore introduce \emph{Budgeted PDI Attainment} (BPA), a normalized,
anytime measure for this purpose.  Let $i$ index an Item Placement (IP)
instance, $m$ index a tuning method, and $B$ denote the number of solver
calls made for that instance.  The incumbent PDI after $B$ calls is
\begin{equation}
 \begin{aligned}
 I_{i,m}(B)
 &= \min_{1 \leq b \leq B} \operatorname{PDI}_{i,m,b},\\
 \operatorname{BPA}_{i,m}(B)
 &= \operatorname{clip}\!\left(
 \frac{PDI_i^{\star}}{I_{i,m}(B)},\,0,\,1\right).
 \end{aligned}
 \label{eq:bpa_definition}
\end{equation}
where $PDI_i^{\star}$ is the frozen best-known PDI for instance $i$,
constructed from the audited result panel, and
$\operatorname{clip}(x,0,1)=\min\{1,\max\{0,x\}\}$.  Since lower PDI is
better, larger BPA is better: $\operatorname{BPA}=1$ means that the
method reaches the current best-known PDI, while
$\operatorname{BPA}=0.8$ means that its incumbent PDI is $1.25$ times the
best-known value.  The normalization is performed separately for every
instance, which makes values comparable across instances with very
different PDI scales.

For methods with repeated runs, we first average BPA over runs for each
instance and then average over the 100 IP instances; consequently, methods
with more repeated runs do not receive extra weight.  The reported
95\% confidence intervals are obtained with 5,000 bootstrap resamples of
instances.  The budget axis counts solver calls rather than wall-clock
seconds.  If a run stops before a plotted budget, its incumbent is carried
forward, and the corresponding observed-evaluation coverage is reported
separately.  All five trajectory methods (GRIMIP, SMAC-I, ifBO, TuRBO, and
LLAMBO) have complete coverage through the strict common budget
$B=8$; GPTT has endpoint-only data and is therefore not included in
anytime BPA curves or BPA--AUC.

To summarize early-budget efficiency without depending on a single budget
point, we also report the discrete area under the BPA curve,
\begin{equation}
 \operatorname{BPA\text{-}AUC}@B
 = \frac{1}{B}\sum_{b=1}^{B}\operatorname{BPA}(b).
 \label{eq:bpa_auc}
\end{equation}
Higher BPA--AUC indicates that a method reaches high attainment earlier.
Following the requested scope, we retain this scalar efficiency summary but
omit a separate ``Budget efficiency ranking'' plot and the per-instance
pairwise win-score matrix.

\subsection{Fixed-Budget Results at $B=8$} \label{sec:bpa_b8}

Table~\ref{tab:bpa_b8} reports the principal fixed-budget comparison.  At
$B=8$, GRIMIP obtains the highest mean BPA (0.7844), followed by LLAMBO
(0.7337), ifBO (0.7296), TuRBO (0.7059), and SMAC-I (0.6720).  The
GRIMIP--LLAMBO gap is $0.0506$ BPA points, with a paired
instance-bootstrap 95\% confidence interval for
LLAMBO$-$GRIMIP of $[-0.0862,-0.0181]$.  GRIMIP also reaches
$\operatorname{BPA}\geq 0.80$ on 56\% of the instances, compared with
45\% for LLAMBO, and reaches the best-known panel exactly on 7\% of
instances, compared with 5\% for LLAMBO.

\begin{table*}[t]
    \centering
    \scriptsize
    \setlength{\tabcolsep}{3pt}
    \caption{Fixed-budget BPA comparison on 100 IP instances at
    $B=8$. Values are means over instances; brackets show 95\%
    instance-bootstrap confidence intervals.  $\Delta$ is the mean BPA
    difference relative to GRIMIP (method $-$ GRIMIP).}
    \label{tab:bpa_b8}
    \begin{tabularx}{\textwidth}{@{}l *{6}{C}@{}}
    \toprule
    \textbf{Method} &
    \shortstack{\textbf{BPA@8}\\\textbf{(95\% CI)}} &
    \shortstack{\textbf{BPA}\\\textbf{$\geq$0.80}} &
    \shortstack{\textbf{BPA}\\\textbf{$\geq$0.90}} &
    \shortstack{\textbf{BPA}\\\textbf{$=1.00$}} &
    \shortstack{\textbf{BPA--AUC@8}\\\textbf{(95\% CI)}} &
    \shortstack{\textbf{$\Delta$ BPA}\\\textbf{vs.\ GRIMIP}} \\
    \midrule
    \rowcolor{grimip_color!20}
    GRIMIP & \textbf{0.7844 [0.7539, 0.8149]} & \textbf{56\%} & \textbf{22\%} & \textbf{7\%} &
    \textbf{0.6634 [0.6316, 0.6940]} & \textbf{0.0000} \\
    LLAMBO & 0.7337 [0.6969, 0.7691] & 45\% & 20\% & 5\% &
    0.6638 [0.6292, 0.6973] & -0.0506 \\
    ifBO & 0.7296 [0.6923, 0.7661] & 40\% & 19\% & 0\% &
    0.6512 [0.6163, 0.6855] & -0.0547 \\
    TuRBO & 0.7059 [0.6693, 0.7410] & 33\% & 13\% & 0\% &
    0.6273 [0.5920, 0.6614] & -0.0785 \\
    SMAC-I & 0.6720 [0.6369, 0.7069] & 29\% & 8\% & 2\% &
    0.5775 [0.5423, 0.6116] & -0.1123 \\
    \bottomrule
    \end{tabularx}
\end{table*}

The BPA--AUC values are nearly tied for GRIMIP (0.6634) and LLAMBO
(0.6638); the paired 95\% interval for LLAMBO$-$GRIMIP is
$[-0.0186,0.0194]$.  This near tie should be interpreted in light of the
protocol: GRIMIP's first five evaluations are its initialization population
and already provide a strong early gain.  Hence, BPA--AUC@8 aggregates that
initialization phase with only three subsequent sequential decisions; it does
not establish an early-search advantage for LLAMBO.  GRIMIP has the stronger
attainment at the common-budget endpoint.

\subsection{Attainment Profiles and Anytime Behavior} \label{sec:bpa_profiles}

\begin{figure}[t]
    \centering
    \includegraphics[width=\linewidth]{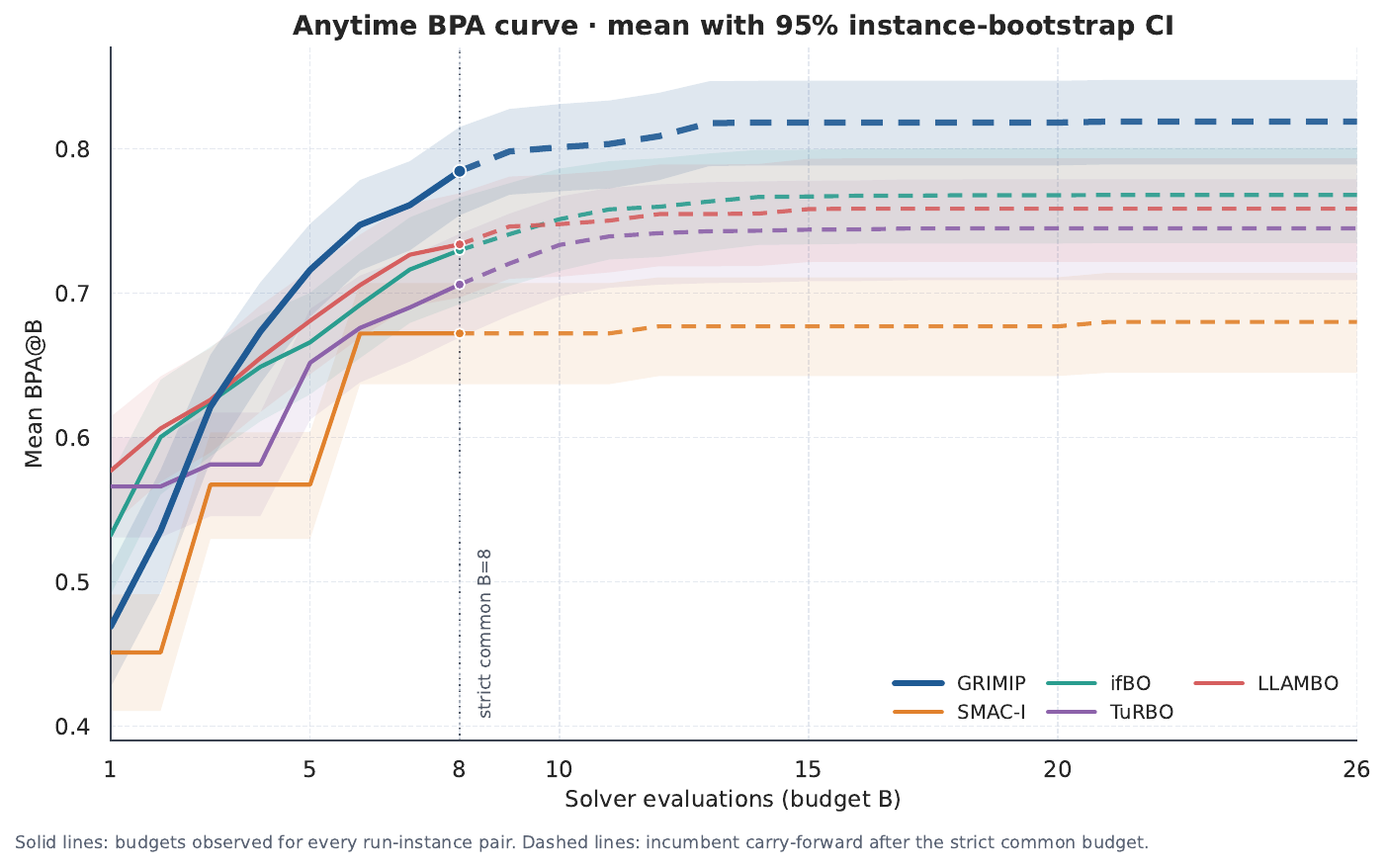}
    \caption{Anytime BPA on the IP dataset. Curves show the mean
    per-instance BPA and 95\% instance-bootstrap intervals.  The vertical
    marker denotes the strict common budget $B=8$; dashed portions use
    incumbent carry-forward after a method has stopped on some
    instances.}
    \label{fig:bpa_anytime}
\end{figure}

Figure~\ref{fig:bpa_anytime} shows the mean anytime BPA curve with
instance-bootstrap intervals.  Solid curves correspond to the
strictly-comparable region through $B=8$; beyond that point, the curves
use incumbent carry-forward because the methods stop at different
budgets.  GRIMIP remains above the other trajectory methods at the common
endpoint.  The first five GRIMIP evaluations constitute its initialization
population, so the early segment should not be interpreted as an isolated
comparison of sequential search behavior.

\begin{figure}[t]
    \centering
    \includegraphics[width=\linewidth]{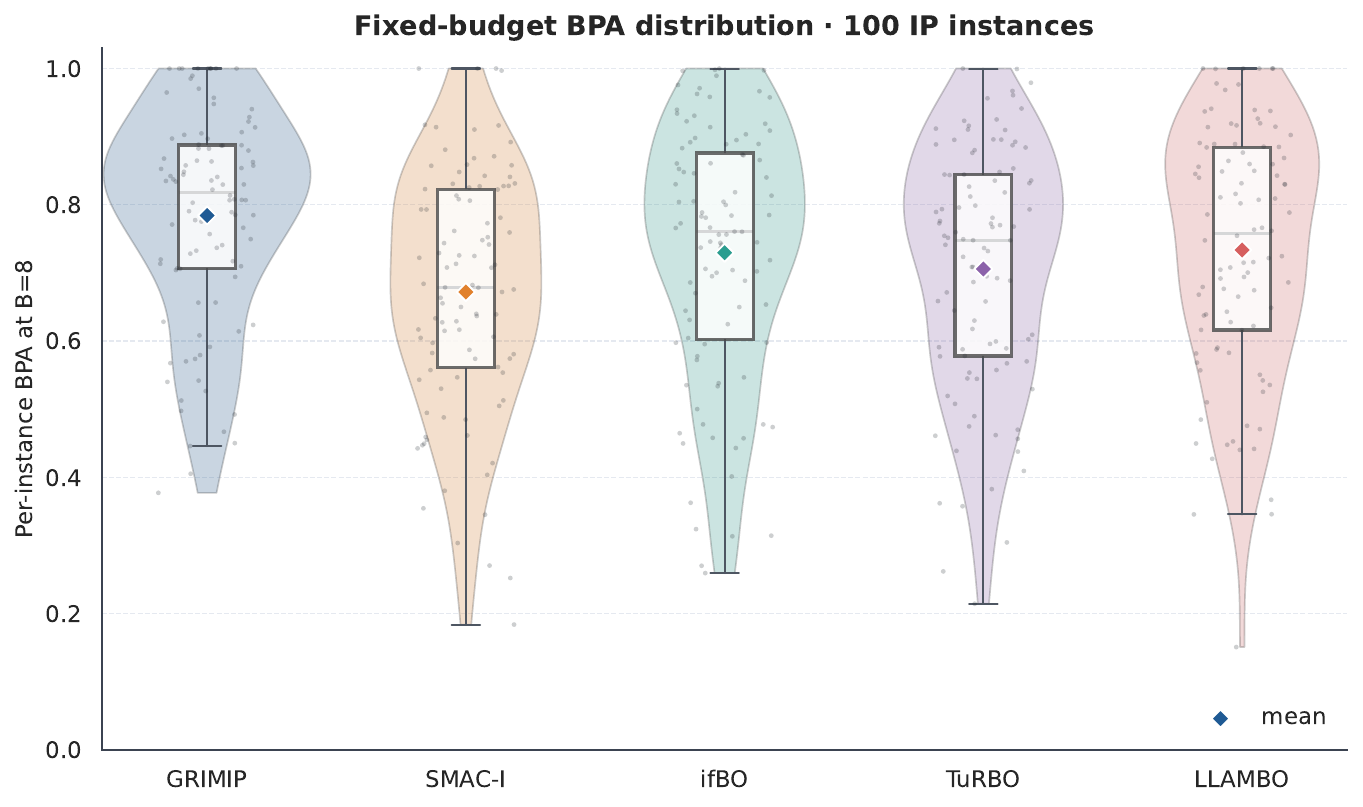}
    \caption{Per-instance BPA distribution at $B=8$. Violin
    shapes show the distribution over 100 IP instances; white boxes show
    the interquartile range and diamonds mark the mean.}
    \label{fig:bpa_distribution}
\end{figure}

The distribution of per-instance BPA at the common budget is shown in
Figure~\ref{fig:bpa_distribution}.  In addition to its higher mean,
GRIMIP has a visibly higher central mass than SMAC-I and TuRBO, while the
spread across instances confirms that the mean alone does not describe
the difficulty of the full benchmark.

\begin{figure}[t]
    \centering
    \includegraphics[width=\linewidth]{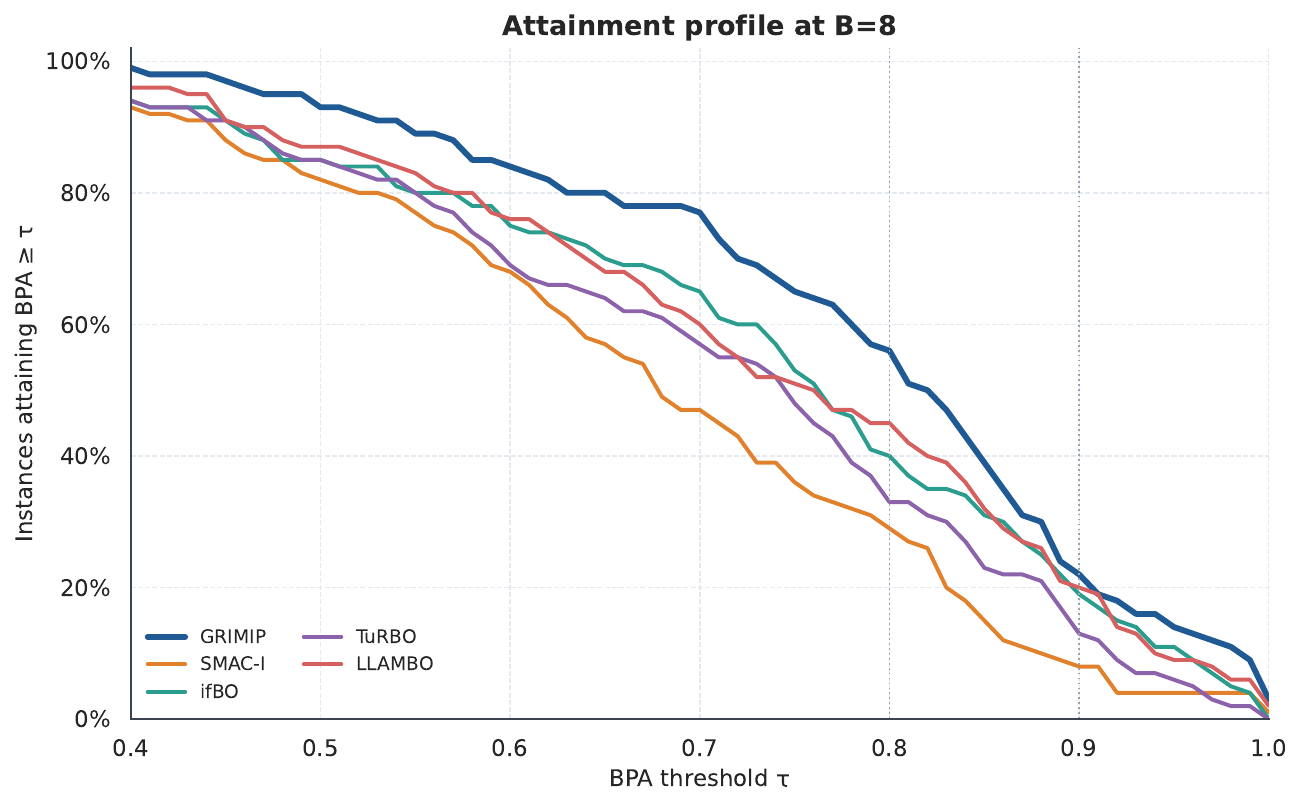}
    \caption{Threshold-attainment profile at $B=8$. Each curve
    reports the fraction of IP instances attaining at least threshold
    $\tau$; higher curves are better.}
    \label{fig:bpa_attainment_profile}
\end{figure}

Finally, Figure~\ref{fig:bpa_attainment_profile} reports the full
threshold-attainment profile at $B=8$: for a threshold $\tau$, the curve
gives the fraction of instances with $\operatorname{BPA}\geq\tau$.
This view makes the result robust to the choice of a single threshold and
shows that GRIMIP maintains the highest attainment over most of the
useful BPA range.

\FloatBarrier
\section{Warm-starting Features} \label{ws_features}

Warm-starting uses instance-level features to construct an informative initial configuration set before the Bayesian optimization loop. We collect static features directly from the MIP model and dynamic features from a 30-second preliminary run with Gurobi's default settings. The exact feature sets used by WS are listed below.

\begin{table}[!t]
\centering
\scriptsize
\setlength{\tabcolsep}{2pt}
\caption{Static Features of MIP Instances}
\label{tab:static_features}
\begin{tabularx}{\columnwidth}{@{}p{0.22\columnwidth}p{0.31\columnwidth}X@{}}
\toprule
\textbf{Category} & \textbf{Feature} & \textbf{Description} \\
\midrule
\multirow{6}{=}{Variable-related} & Total number of variables & The total scale of decision variables in the problem. \\
 & Number of integer variables & The count of variables restricted to integer values. \\
 & Number of binary variables & The count of 0-1 variables, a subset of integer variables. \\
 & Number of continuous variables & The count of variables that can take any real value. \\
 & \multirow{2}{=}{Ratios of variable types} & Proportion of binary variables to total variables. \\
 & & Proportion of integer (incl. binary) variables to total. \\
\midrule
\multirow{7}{=}{Constraint-related} & Total number of constraints & The total number of constraint equations in the problem. \\
 & \multirow{3}{=}{Constraint types} & Number of equality (=) constraints. \\
 & & Number of less-than-or-equal-to ($\leq$) constraints. \\
 & & Number of greater-than-or-equal-to ($\geq$) constraints. \\
\cmidrule(l){2-3}
 & \multirow{3}{=}{Constraint matrix (A) features} & Number of non-zero elements and matrix density. \\
 & & Statistics on non-zeros per row/column (mean, std, etc.). \\
 & & Statistics on coefficient values (mean, variance, etc.). \\
\midrule
\multirow{2}{=}{Objective Function} & Number of non-zero coefficients & Number of variables directly impacting the objective value. \\
 & Statistics on coefficients & Distribution (mean, variance, etc.) of objective coefficients. \\
\bottomrule
\end{tabularx}
\end{table}

\begin{table}[!t]
\centering
\scriptsize
\setlength{\tabcolsep}{2pt}
\caption{Dynamic Features of MIP Instances (from a 30-second run)}
\label{tab:dynamic_features}
\begin{tabularx}{\columnwidth}{@{}p{0.22\columnwidth}p{0.31\columnwidth}X@{}}
\toprule
\textbf{Category} & \textbf{Feature} & \textbf{Description} \\
\midrule
\multirow{4}{=}{Root Node / LP Relaxation} & Initial dual gap & Gap between root LP objective and initial feasible solution. \\
 & Time to solve root relaxation & Time required to solve the initial LP relaxation. \\
 & Objective value of root relaxation & The initial value of the dual bound. \\
 & Number of simplex iterations & Simplex iterations needed for the root relaxation. \\
\midrule
\multirow{3}{=}{Branch-and-Bound Tree} & Number of processed nodes & Total B\&B nodes explored within the time limit. \\
 & Number of open nodes left & Unprocessed nodes in the B\&B tree at the time limit. \\
 & Maximum depth of the tree & The greatest branching depth reached during the run. \\
\midrule
\multirow{5}{=}{Cutting Planes \& Presolve} & \multirow{2}{=}{Types \& counts of cuts} & Number of specific cuts applied (Gomory, MIR, Cover, etc.). \\
 & & Total number of all generated cutting planes. \\
\cmidrule(l){2-3}
 & \multirow{3}{=}{Presolve effectiveness} & Number of variables removed by presolve. \\
 & & Number of constraints removed by presolve. \\
 & & Time spent in the presolve phase. \\
\midrule
\multirow{3}{=}{Heuristics \& Solutions} & Number of feasible solutions & Count of integer-feasible solutions found. \\
 & Objective of first solution & Quality of the first solution found by heuristics. \\
 & Objective of best solution & Quality of the best solution (primal bound) found. \\
\bottomrule
\end{tabularx}
\end{table}

\FloatBarrier
\section{Evaluation Metrics} \label{metric}

\subsection{Primal-Dual Integral}
\begin{figure}[t]
    \centering
    \includegraphics[width=1\linewidth]{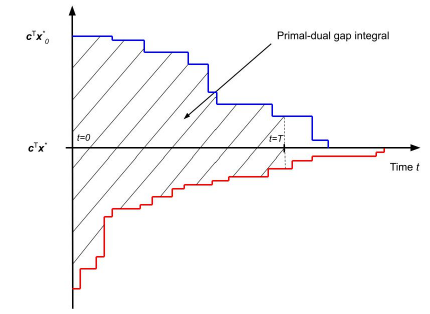}
    \caption{PDI Illustration}
    \label{fig:placeholder}
\end{figure}
When evaluating the performance of a MIP solver, relying solely on total solution time is insufficient, especially for difficult instances that are not solved to optimality within a time limit $T$. We therefore use the \textbf{Primal-Dual Integral (PDI)} to summarize how rapidly the solver closes its primal-dual gap over the full scoring horizon.

Let $p(t)$ denote the objective value of the best incumbent solution and let $d(t)$ denote the best dual bound. At every valid observation, we compute the clipped gap
\[
g(p,d)=
\begin{cases}
\displaystyle \min\!\left\{1,\frac{\lvert p-d\rvert}{\lvert p\rvert}\right\},
& \lvert p\rvert>10^{-10},\\[6pt]
\displaystyle \min\!\left\{1,\lvert p-d\rvert\right\},
& \lvert p\rvert\leq 10^{-10}.
\end{cases}
\]
Thus, the integrand is a dimensionless relative gap clipped at one, rather than the absolute difference between the two objective bounds. The absolute-difference fallback near a zero incumbent avoids division by a numerically negligible value. If the primal and dual bounds straddle zero, the relative gap will typically exceed one and is therefore clipped to one.

For valid observations at times $0=t_0<t_1<\cdots<t_N=T$, with $g_i=g\bigl(p(t_i),d(t_i)\bigr)$, we use the left-rectangle rule
\[
\operatorname{PDI}_T=\sum_{i=0}^{N-1}g_i\,(t_{i+1}-t_i).
\]
The metric is not divided by $T$. Consequently, its unit is seconds, and its usual range is $[0,T]$; in our 300-second evaluations, this is ordinarily $[0,300]$. Lower values are better: improvements to either the incumbent or the dual bound reduce the subsequent contribution to the integral. A value of zero requires the recorded gap to be zero throughout the scoring horizon, whereas closing the gap at a positive time still incurs the area accumulated beforehand.

We use the following conventions for incomplete or degenerate trajectories. The gap is initialized to one at $t=0$. Until both a feasible incumbent and a valid dual bound are available, no new observation is added, so the preceding gap---initially one---persists. If no feasible solution is found during the entire run, the score is $T$. If a feasible solution exists but there are no valid intermediate observations, the score is defined as $T$ times the final solver-reported relative gap, irrespective of the actual solution time. For a normally recorded trajectory, the last interval is extended to $T$ using the final gap. Hence, when a run terminates early, its terminal gap persists from termination to $T$; if optimality has been proved, this terminal gap is zero.

Two minor edge cases can place the score outside its usual range. The terminal solver-reported gap is used without an additional clipping step, so a degenerate trajectory with a reported gap above one can yield a score greater than $T$. In addition, observation times are not clipped to the scoring horizon; a final observation marginally later than $T$ can make the numerical integral slightly exceed the nominal time limit.

\subsection{Solving Time}
In the domain of mathematical optimization, solution speed is a metric of paramount importance. We define Solving Time as the wall-clock time required by the Gurobi solver to find a proven optimal solution for a given problem instance. The solver's execution for any single run terminates under one of two conditions:

\begin{itemize}
    \item Optimality: The solver finds a solution and successfully proves its optimality.
    \item Time Limit: The solver exhausts a predefined computational budget before optimality is proven.
\end{itemize}

For our evaluation, particularly on the MIK and CORAL datasets, this metric is the primary performance indicator. The objective of any tuning method is to find a parameter configuration that minimizes this solving time.

\FloatBarrier
\section{Baseline Methods} \label{baseline}

To comprehensively evaluate efficacy, we compare GRIMIP against a suite of established baselines. To ensure a strictly fair comparison regarding the search landscape, \textbf{all baseline methods} are configured to explore the exact same hyperparameter space $\Theta$, consisting of \textbf{6 high-impact parameters} (detailed in Section \ref{sec:random-search}). This focused selection reflects realistic operational constraints, ensuring that the optimization process remains tractable and avoids the risk of the \textit{curse of dimensionality} associated with navigating high-dimensional configuration spaces.

\subsection{Random Search (Random)}\label{sec:random-search}

To rigorously assess the necessity of intelligent optimization strategies, we include \textbf{Random Search} as a fundamental baseline. It explores the hyperparameter space without any learning mechanism or historical guidance.

\paragraph{Configuration Space}
To eliminate discrepancies in the accessible solution space, we define a standardized configuration space $\Theta$ used by \textbf{all baselines} (including SMAC-I, SMAC-P, LLAMBO, ifBO, and TuRBO). Instead of exploring the full set of available Gurobi parameters, we focus on the \textbf{6 most influential parameters} to prevent high-dimensional sparsity issues. The space $\Theta$ comprises:
\begin{itemize}
    \item \textbf{MIPFocus} (int, $0-3$): Controls the high-level solution strategy (e.g., focusing on feasibility vs. optimality).
    \item \textbf{Heuristics} (float, $0.0-1.0$): Determines the percentage of time spent on primal heuristics.
    \item \textbf{Cuts} (int, $-1$ to $3$): Global aggressiveness of cutting plane generation.
    \item \textbf{Presolve} (int, $-1$ to $2$): Controls the level of presolving applied before the branch-and-bound search.
    \item \textbf{Method} (int, $-1$ to $2$): Selects the specific algorithm used to solve the root relaxation (e.g., primal/dual simplex, barrier).
    \item \textbf{VarBranch} (int, $-1$ to $3$): Determines the variable branching strategy within the search tree.
\end{itemize}

\paragraph{Setup}
The evaluation budget is coupled to GRIMIP's computational effort. For each instance, Random Search samples $K = N_{iter} + N_{init}$ configurations uniformly at random from this 6-parameter space. The configuration yielding the minimum Primal-Dual Integral (PDI) is selected as the final solution.

\subsection{SMAC-Population (SMAC-P)}\label{sec:smac-population}

\textbf{SMAC-P} serves as a global baseline, seeking a single \textit{``one-size-fits-all''} configuration that minimizes the average PDI across the entire instance distribution.

\paragraph{Setup}
Operating within the identical \textbf{6-parameter space} $\Theta$, SMAC-P is constrained by a strict budget of \textbf{500 trials} or \textbf{7200 seconds}. In each iteration, the optimizer evaluates a candidate configuration on a stochastic subset of 10 instances (single-threaded). The process terminates when the budget is exhausted, returning the incumbent configuration with the best average performance.

\subsection{SMAC-Individual (SMAC-I)} \label{sec:smac-individual}

\textbf{SMAC-I} acts as the rigorous per-instance baseline, executing an independent configuration process for each specific MIP instance.

\paragraph{Setup}
SMAC-I seeks the optimal configuration $\theta^* \in \Theta$ to minimize the instance-specific PDI. It explicitly operates within the same 6-parameter search space $\Theta$ defined for all other baselines. To ensure a strictly fair comparison, the total wall-clock tuning budget for each instance is aligned with the time consumed by GRIMIP. Additionally, all evaluations are conducted on a single thread to eliminate variance arising from parallelism.

\subsection{Gurobi Parameter Tuning Tool (GPTT)} \label{sec:gptt}

We incorporate \textbf{GPTT}, Gurobi's native automated tuning utility.

\paragraph{Setup}
GPTT is configured to operate within the identical parameter space $\Theta$ as the other methods, restricted to the 6 high-impact parameters. We set \texttt{TuneTrials=3} to perform 3 independent runs per candidate. To ensure fairness, the tuning budget passed to GPTT ($T_{tune}^{(i)}$) is dynamically derived from GRIMIP's total wall-clock time ($T_{total}^{(i)}$), reserving a fixed duration ($T_{eval} = 300\text{s}$) for the final validation run:
\[
    T_{tune}^{(i)} = \max\left( T_{total}^{(i)} - T_{eval}, \quad 300\text{s} \right)
\]

\subsection{LLAMBO}

\textbf{LLAMBO} \citep{liu2024llmbo} is an LLM-driven framework using Monte Carlo sampling for performance and uncertainty estimation. While the original framework is effective on low-dimensional spaces, we ensure a fair comparison by configuring LLAMBO to explore the exact same \textbf{6-parameter space} $\Theta$ as the other baselines. This dimensionality aligns well with LLAMBO's typical operating range, avoiding the inefficiencies often observed in higher-dimensional tasks. We set the Monte Carlo sampling frequency to 5 independent queries per candidate. The budget is aligned with GRIMIP (at most 50 solver evaluations, an initial population of 5, and a patience of 4), ensuring the comparison focuses on the search methodology rather than the search space size.

\subsection{ifBO} \label{sec:ifbo}

\textbf{ifBO} \citep{rakotoarison2024ifbo} is a recently proposed in-context freeze-thaw Bayesian optimization method. It replaces the hand-crafted Gaussian-process surrogate of classical freeze-thaw BO with a pre-trained transformer (FT-PFN) that acts as a learning-curve prior, enabling the optimizer to adaptively decide whether to continue evaluating a promising configuration or to discard it in favour of a new candidate. This makes ifBO a particularly relevant baseline for solver tuning, where configuration evaluations are expensive and can in principle be terminated early based on partial convergence information.

\paragraph{Setup}
ifBO operates within the identical \textbf{6-parameter space} $\Theta$ used by all other baselines, and is \emph{not} allowed to access any parameters outside this set. To guarantee a strictly fair comparison, the total tuning effort for each instance is capped by the same \textbf{wall-clock time} consumed by GRIMIP on that instance (rather than by a fixed number of trials), and all evaluations run single-threaded. The initial design, acquisition function, and FT-PFN surrogate are used as released by the authors; only the budget, search space, and evaluation oracle (Gurobi with single-thread PDI evaluation) are standardised for our experiments.

\subsection{TuRBO} \label{sec:turbo}

\textbf{TuRBO} \citep{eriksson2019turbo} is a trust-region Bayesian optimization method designed to scale classical BO beyond low-dimensional problems. Rather than fitting a single global Gaussian-process surrogate, TuRBO maintains one or more local GP surrogates inside shrinking/expanding trust regions that are restarted when they collapse. We include it as a strong \emph{non-LLM}, per-instance BO baseline that is known to be competitive on high-dimensional black-box tuning.

\paragraph{Setup}
TuRBO is restricted to the same \textbf{fixed 6-parameter space} $\Theta$ as the other baselines; no additional Gurobi parameters are exposed to it. As with SMAC-I and ifBO, the per-instance tuning budget is set by matching GRIMIP's \textbf{wall-clock time} on that instance, and all solver evaluations are executed single-threaded. We use a single trust region (TuRBO-1) with the default initial design size, trust-region expansion/contraction thresholds, and acquisition rule from the original implementation; only the search space and budget are adapted to our benchmark.

\FloatBarrier
\section{Algorithms}

Algorithm~\ref{alg:grimip} clearly demonstrates the four core stages of GRIMIP: feature extraction, automatic spatial selection (ASS), Warmstart, and LLM-driven Bayesian optimization loops.

\vspace{5pt}

\begin{algorithm}[t]
\caption{GRIMIP: General Reasoning for Instance-specific MIP Configuration}
\label{alg:grimip}
\begin{algorithmic}[1]
\REQUIRE MIP instance $P$, LLM, Budget $N_{iter}$, Time Limit $T_{limit}$
\ENSURE Optimized configuration $c^*$

\STATE \textbf{Feature Extraction:}
\STATE $\mathcal{F}_{static} \leftarrow \text{ExtractStatic}(P)$
\STATE $\mathcal{F}_{dynamic} \leftarrow \text{RunTrial}(P, 30s)$
\STATE $\mathcal{F}_P \leftarrow \mathcal{F}_{static} \cup \mathcal{F}_{dynamic}$

\STATE \textbf{Automated Space Selection (ASS):}
\STATE $\mathcal{S} \leftarrow \text{LLM}(\text{Prompt}_{ASS}, \mathcal{F}_P)$

\STATE \textbf{Warmstart:}
\STATE $\mathcal{P}_{init} \leftarrow \text{LLM}(\text{Prompt}_{WS}, \mathcal{F}_P, \mathcal{S})$
\STATE Evaluate $\mathcal{P}_{init}$ to initialize history $\mathcal{D}$

\STATE \textbf{Bayesian Optimization Loop:}
\WHILE{$n < N_{iter}$}
    \STATE \textit{// 1. Candidate Generation}
    \STATE $\{\tilde{c}_k\}_{k=1}^K \leftarrow \text{LLM}(\text{Prompt}_{Gen}, \mathcal{D}, \mathcal{S})$
    
    \STATE \textit{// 2. Surrogate Modeling (Joint Prediction)}
    \FOR{$k=1$ to $K$}
        \STATE $(\mu_k, \sigma_k) \leftarrow \text{LLM}(\text{Prompt}_{Surrogate}, \tilde{c}_k, \mathcal{D})$
        \STATE $\alpha_k \leftarrow \text{AcquisitionFunc}(\mu_k, \sigma_k)$
    \ENDFOR
    
    \STATE \textit{// 3. Evaluation \& Update}  
    \STATE $c_{new} \leftarrow \argmax_k \alpha_k$
    \STATE $y_{new} \leftarrow \text{Solver}(P, c_{new}, T_{limit})$
    \STATE $\mathcal{D} \leftarrow \mathcal{D} \cup \{(c_{new}, y_{new})\}$
    
    \STATE \textit{// 4. Dynamic Space Refinement (Optional)}
    \STATE $\mathcal{S} \leftarrow \text{LLM}(\text{Prompt}_{Update}, \mathcal{D})$
\ENDWHILE
\STATE \textbf{return} Best $c^*$ in $\mathcal{D}$
\end{algorithmic}
\end{algorithm}

\FloatBarrier
\section{Ablation Experiments}

\subsection{The Role of Uncertainty} \label{uncertainty}
\begin{figure}
    \centering
    \includegraphics[width=0.8\linewidth]{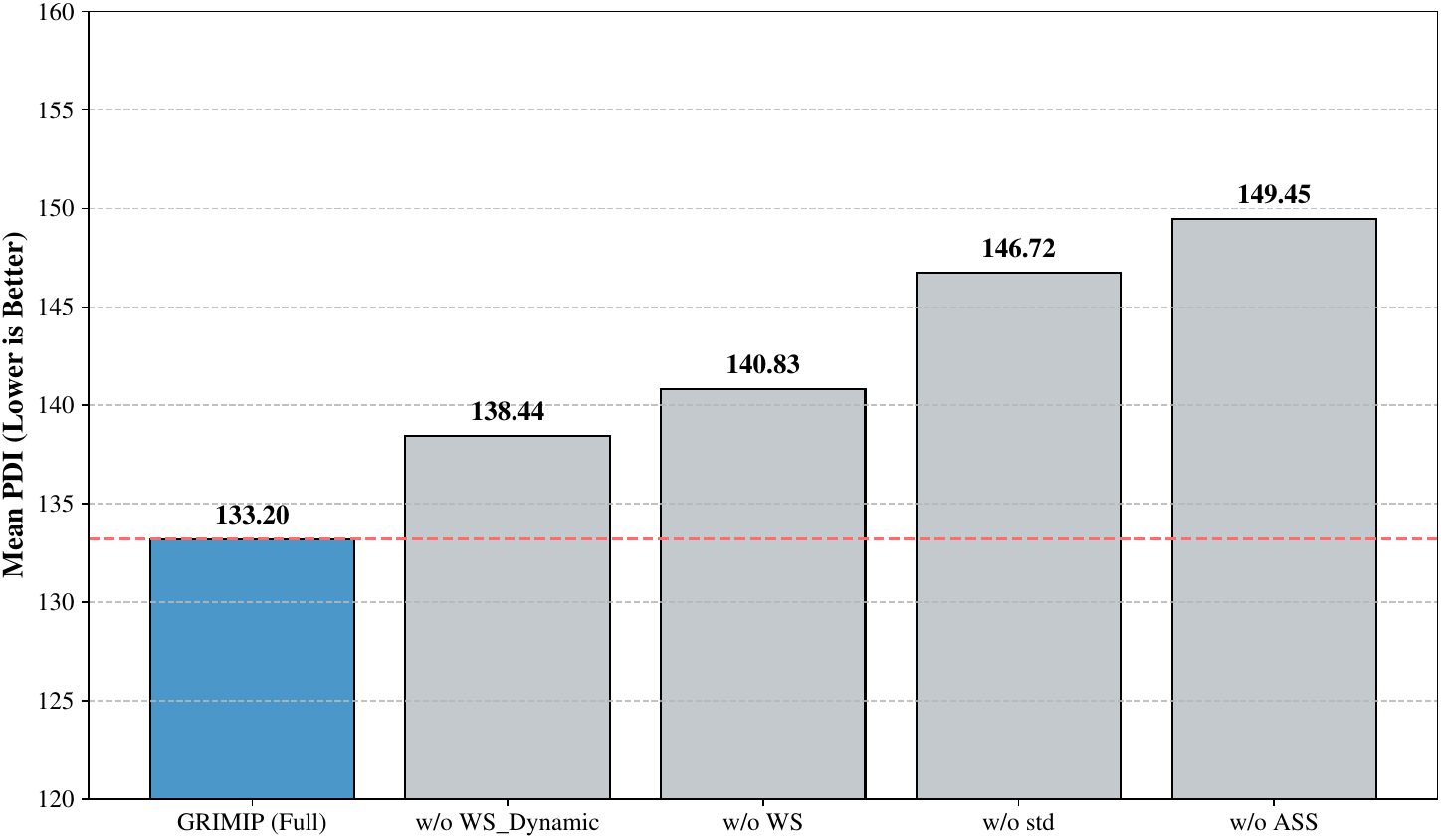}
    \caption{Ablation study of GRIMIP components on the IP dataset. The bar chart compares the Mean Primal-Dual Integral (PDI) of the full framework against variants with specific modules removed. The \emph{w/o ASS} bar denotes the fixed-six-parameter variant; the complementary fixed-16-parameter result is reported in the main ablation table.}
    \label{fig:abl-bar}
\end{figure}

To quantify the critical role of uncertainty modeling (i.e., predicting the standard deviation \texttt{std}) within our framework, we conducted a rigorous ablation study, comparing two versions of our method:
\textcircled{\scriptsize 1}~GRIMIP (Full Version): Leverages a single prompt call for the LLM to simultaneously predict both the performance mean and its uncertainty (\texttt{std}). \textcircled{\scriptsize 2}~GRIMIP \textbf{w/o std }(Simplified Version): As a control group, this version adopts an approach following the LLAMBO framework \citep{liu2024llmbo} to indirectly acquire uncertainty and performance. Specifically, it only prompts the LLM to predict the performance. To calculate the \texttt{std} required for the Bayesian optimization acquisition function, this method repeatedly calls the LLM for the same candidate point (5 times) to obtain multiple independent performance predictions. The \texttt{std} of these predictions is then calculated and used as the uncertainty estimate. As shown in Figure~\ref{fig:abl-bar} in terms of solution quality, the full GRIMIP version demonstrated superior performance with a mean PDI of \textbf{133.20} compared to \textbf{146.72} for GRIMIP w/o std. This improvement showcases the effectiveness of direct uncertainty modeling in achieving better optimization results. The full GRIMIP version also demonstrated slightly better efficiency, requiring an average of \textbf{11.30} evaluations compared to \textbf{11.52} evaluations for the simplified version, highlighting the benefits of joint prediction of both mean and uncertainty in a single LLM call.

The results confirm that direct uncertainty modeling provides more informed estimates that enable a superior balance between ``exploration" and ``exploitation" in the Bayesian optimization process.

\subsection{The Role of Warm-starting} 
To isolate and quantify the contribution of our instance-aware intelligent initialization, we conducted an ablation study comparing three versions: the full GRIMIP framework (which uses both static and dynamic features for warm-starting), GRIMIP \textbf{w/o WS\_Dynamic} (which uses only static features for warm-starting), and GRIMIP \textbf{w/o WS} (which uses no warm-starting phase at all).

The results in Figure~\ref{fig:abl-bar} highlight the critical role of this stage. The full GRIMIP framework achieved the best average PDI of \textbf{133.20}. This represents a notable improvement over the static-only warmstart (\textbf{138.44} PDI) and is significantly better than having no warmstart (\textbf{140.83} PDI). This demonstrates that warm-starting is effective, and that incorporating dynamic features provides an additional performance boost.

This performance boost confirms that the warm-starting phase effectively addresses the ``cold-start" problem. Furthermore, the comparison between GRIMIP and GRIMIP w/o WS\_Dynamic demonstrates that leveraging \textit{dynamic} features, in addition to static ones, provides a distinct and valuable advantage. This allows the framework to provide a more informed and strategically advantageous starting point, leading to a more robust and efficient search.

\subsection{The Role of ASS}
To validate the effectiveness of the ASS module, we compare the full GRIMIP framework against two variants that disable instance-specific space selection. Full GRIMIP uses the LLM to analyze instance features and dynamically select a low-dimensional (\(k=6\)) parameter subset for each problem. \emph{GRIMIP w/o ASS (fixed 6 params)} instead uses a predefined ``one-size-fits-all" space composed of MIPFocus, Heuristics, Cuts, Presolve, Method, and VarBranch, whereas \emph{GRIMIP w/o ASS (16 params)} operates over a broader fixed 16-parameter space.

The results clearly demonstrate the advantage of ASS. Full GRIMIP achieves the lowest mean PDI of 133.20, while the fixed-six and fixed-16 variants obtain mean PDIs of 149.45 and 153.98, respectively. Thus, simply exposing more parameters without instance-aware selection does not recover the benefit of ASS.

The fixed-six variant uses fewer evaluations on average than full GRIMIP (10.59 versus 11.30), suggesting premature convergence in a rigid space that may omit influential controls. Conversely, the fixed-16 variant uses more evaluations (11.68) but performs even worse, indicating that the broader unpruned space diffuses the limited search budget. Together, these results show that ASS is valuable not merely because it reduces dimensionality, but because it focuses the search on parameters relevant to each instance.

As an additional optimizer-controlled experiment, we run standard SMAC-I in the instance-specific six-parameter spaces selected by ASS while retaining the original single-thread setting and per-instance budget (denoted \textbf{SMAC-I+ASS}). It obtains a mean PDI of \textbf{149.77} with \textbf{11.68} evaluations on average, compared with \textbf{152.28} for the original fixed-space SMAC-I.

\textbf{Remark.} Holding the optimizer and budget fixed, this comparison indicates that the ASS-selected search space is effective.

\bibliography{references}

\end{document}